\documentclass[acmsmall,screen]{acmart}

\AtBeginDocument{%
  \providecommand\BibTeX{{%
    \normalfont B\kern-0.5em{\scshape i\kern-0.25em b}\kern-0.8em\TeX}}}

\setcopyright{rightsretained}

\acmJournal{CSUR}
\acmYear{2023}
\acmVolume{00}
\acmNumber{JA}
\acmArticle{00}
\acmMonth{05}

\acmPrice{15.00}
\acmDOI{10.1145/3597199}

\newcommand{\Yhat}{\widehat{Y}}
\newcommand{\yhat}{\hat{y}}

\usepackage{amsmath}
\usepackage{amssymb}
\usepackage{amsthm}
\usepackage{bm}  
\usepackage{centernot}
\usepackage{color}
\usepackage{enumitem}  
\usepackage{float}  
\usepackage{graphicx}  
\usepackage{mathtools}
\usepackage{microtype}  
\usepackage{multirow}  
\usepackage{subcaption}  
\usepackage{textcomp}  
\usepackage{verbatim}  

\theoremstyle{plain}
\newtheorem{theorem}{Theorem}[section]

\theoremstyle{definition}
\newtheorem{definition}[theorem]{Definition}

\theoremstyle{remark}

\newcommand\indep{\protect\mathpalette{\protect\independent}{\perp}}
\def\independent#1#2{\mathrel{\rlap{$#1#2$}\mkern2mu{#1#2}}}
\newcommand\nindep{\centernot\indep}

\newcommand{\Fbf}{\mathbf{F}}

\newcommand{\Acal}{\mathcal{A}}
\newcommand{\Bcal}{\mathcal{B}}
\newcommand{\Ccal}{\mathcal{C}}

\newcommand{\Gcal}{\mathcal{G}}

\newcommand{\Xcal}{\mathcal{X}}
\newcommand{\Ycal}{\mathcal{Y}}

\usepackage[disable,textwidth=1.3in,textsize=footnotesize]{todonotes}

\usepackage{xcolor}
\definecolor{FillColorCBBlue}{HTML}{EDF9FF}
\definecolor{FillColorCBGreen}{HTML}{EDFFFA}
\definecolor{FillColorCBOrange}{HTML}{FFF5ED}
\definecolor{FillColorCBPurple}{HTML}{FFF8FD}
\definecolor{FillColorCBYellow}{HTML}{FFFEF1}

\definecolor{CustomIvory}{HTML}{FCFCED}

\definecolor{FontColorBlue}{HTML}{0433FF}
\definecolor{FontColorGreen}{HTML}{009051}
\definecolor{FontColorOrange}{HTML}{941100}
\definecolor{FontColorPurple}{HTML}{FF2F92}
\definecolor{FontColorYellow}{HTML}{929000}

\definecolor{StrokeColorCBBlue}{HTML}{0173B2}
\definecolor{StrokeColorCBGreen}{HTML}{029E73}
\definecolor{StrokeColorCBOrange}{HTML}{D55E00}
\definecolor{StrokeColorCBPurple}{HTML}{CC78BC}
\definecolor{StrokeColorCBYellow}{HTML}{C6BD2B}

\definecolor{RevisionBorder}{HTML}{D1E7FF}
\definecolor{RevisionHighlight}{HTML}{FEECB4}
\definecolor{RevisionText}{HTML}{000000}  

\newcommand{\reAll}[4]{
    \vspace{#4}
    \todo[backgroundcolor=CustomIvory, bordercolor=RevisionBorder, linecolor=white, textcolor=black, author={\small #1}, caption={#2}]{#3}
    \vspace{-#4}
}

\newcommand{\reReviewerBase}[9]{
    \todo[author=\textcolor{#3}{\small \textit{Reviewer #7:} \textbf{#8}}, backgroundcolor=#1, bordercolor=#2, linecolor=#4, textcolor=#5, tickmarkheight=#6, caption={Re: #8 by \textcolor{#3}{Reviewer #7}}]{#9}
}

\newcommand{\reReviewerTickedVoffset}[7]{
    \vspace{#7}
    \reReviewerBase{#1}{#2}{#3}{#2}{black}{0.1cm}{#4}{#5}{#6}
    \vspace{-#7}
}

\newcommand{\reBlue}[3]{\reReviewerTickedVoffset{FillColorCBBlue}{StrokeColorCBBlue}{FontColorBlue}{One}{#1}{#2}{#3}}
\newcommand{\reOrange}[3]{\reReviewerTickedVoffset{FillColorCBOrange}{StrokeColorCBOrange}{FontColorOrange}{Two}{#1}{#2}{#3}}
\newcommand{\rePurple}[3]{\reReviewerTickedVoffset{FillColorCBPurple}{StrokeColorCBPurple}{FontColorPurple}{Three}{#1}{#2}{#3}}

\begin{document}

\title[What-is and How-to for Fairness in Machine Learning]{What-is and How-to for Fairness in Machine Learning: \\
A Survey, Reflection, and Perspective}

\author{Zeyu Tang}
\email{zeyutang@cmu.edu}
\orcid{0000-0002-4423-4728}
\affiliation{%
    \institution{Carnegie Mellon University}
    \country{United States}
}

\author{Jiji Zhang}
\email{jijizhang@cuhk.edu.hk}
\orcid{0000-0003-0684-2084}
\affiliation{%
    \institution{The Chinese University of Hong Kong}
    \country{Hong Kong}
}

\author{Kun Zhang}
\email{kunz1@cmu.edu}
\orcid{0000-0002-0738-9958}
\affiliation{%
    \institution{Carnegie Mellon University}
    \country{United States}
}

\renewcommand{\shortauthors}{Z. Tang et al.}

\begin{abstract}
    \reAll{Thank Reviewers}{}{
        We are extremely grateful to reviewers and editors for the comments and the time devoted.

        In this revised manuscript, we have carefully considered and incorporated the review comments.
        We use the \textcolor{RevisionText}{blue-colored font} to indicate modifications and/or additions of the material, including:
        \begin{itemize}
            \item lines 143 to 144 in Section \ref{sec:motivating_example},
            \item lines 326 to 339 in Section \ref{sec:causal_modeling},
            \item lines 474 to 498 in Section \ref{sec:notion_IF},
            \item lines 912 to 926 in Section \ref{sec:remark_multi_spectra},
            \item lines 1026 to 1060 in Section \ref{sec:causality_intended_interpretation},
            \item lines 1075 to 1119 in Section \ref{sec:work_against_or_with},
            \item lines 1393 to 1396 in Section \ref{sec:achieve_induced_fairness},
            \item the reposition of Figure \ref{fig:flowchart} into Section \ref{sec:achieve_fairness} (with the line pattern improved to make the figure readable when printed in grayscale).
        \end{itemize}

        We also provide color-coded side notes that correspond to comments/questions by each reviewer (%
            \textbf{\textcolor{FontColorBlue}{Reviewer 1},}
            \textbf{\textcolor{FontColorOrange}{Reviewer 2},}
            \textbf{\textcolor{FontColorPurple}{Reviewer 3}}%
        ) to help locate the related material.

        In light of the review comments, we have included the suggested references and the corresponding analyses, to further improve the completeness of the article.
    }{0em}
    \vspace{1.5em}
    \hrule
    \vspace{0.5em}
    We review and reflect on fairness notions proposed in machine learning literature, and make an attempt to draw connections to arguments in moral and political philosophy, especially theories of justice.
    We survey dynamic fairness inquiries, and further consider the long-term impact induced by current prediction and decision.
    We present a flowchart that encompasses implicit assumptions and expected outcomes of different fairness inquiries on the data generating process, the predicted outcome, and the induced impact, respectively.
    We demonstrate the importance of matching the mission (what kind of fairness to enforce) and the means (which appropriate fairness spectrum to analyze) to fulfill the intended purpose.
\end{abstract}

\ccsdesc[500]{Computing methodologies~Artificial intelligence}
\ccsdesc[500]{Computing methodologies~Machine learning}

\keywords{Algorithmic fairness, causality, bias mitigation, dynamic process, fair machine learning}

\maketitle

\vspace{-0.2em}
\hrule
\vspace{2.2em}

\section{Introduction}\label{sec:introduction}

With the widespread utilization of machine learning models in our daily life, researchers have been thinking about the potential social consequences of the prediction/decision made by algorithms.
To date, there is ample evidence that machine learning models have resulted in discrimination against certain groups of individuals under many circumstances,
for instance, the discrimination in ad delivery when searching for names that can be predictive of the race of an individual \citep{sweeney2013discrimination};
the gender discrimination in job-related ads push \citep{datta2015automated}; stereotypes associated with gender in word embeddings \citep{bolukbasi2016man};
the bias against certain ethnic groups in the assessment of recidivism risk \citep{propublica2016compas,berk2021fairness};
the violation of anti-discrimination law (e.g., Title VII of the 1964 Civil Rights Act) in data mining \citep{barocas2016big}.

In the effort of enforcing fairness in machine learning, various notions as well as techniques to regulate discrimination under different scenarios have been proposed in the literature.
There are multiple different perspectives of fairness analysis.
In terms of the type of relation between variables that is encoded in the fairness criterion, there are associative notions of fairness that are defined in terms of correlation or dependence between variables, e.g., \textit{Demographic Parity} \citep{dwork2012fairness}, \textit{Equalized Odds} \citep{hardt2016equality}, and \textit{Predictive Parity} \citep{dieterich2016compas,chouldechova2017fair,zafar2017fairness};
there are also causal notions of fairness that are defined in terms of causal relation between variables, e.g., \textit{Counterfactual Fairness} \citep{kusner2017counterfactual}, \textit{No Unresolved Discrimination} \citep{kilbertus2017avoiding}, and \textit{Path-specific Counterfactual Fairness} \citep{chiappa2019path,wu2019pc}.
In terms of the scope of application, there are group-level fairness notions, e.g., \textit{Equalized Odds} \citep{hardt2016equality}, \textit{Fairness on Average Causal Effect} \citep{khademi2019fairness}, \textit{Equality of Effort} \citep{huang2020fairness};
there are also individual-level fairness notions, e.g., \textit{Individual Fairness} \citep{dwork2012fairness}, \textit{Counterfactual Fairness} \citep{kusner2017counterfactual}, \textit{Individual Fairness on Hindsight} \citep{gupta2019individual}.
In terms of the technique to eliminate or suppress discrimination, there are \textit{pre-processing} approaches \citep{calders2009building,dwork2012fairness,zemel2013learning,zhang2018mitigating,madras2018learning,creager2019flexibly,zhao2020conditional}, \textit{in-processing} approaches \citep{kamishima2011fairness,perez2017fair,zafar2017fairness,zafar2017constraint,donini2018empirical,song2019learning,mary2019fairness,baharlouei2020renyi,romano2020achieving,wang2020robust,jang2021constructing}, and \textit{post-processing} approaches \citep{hardt2016equality,fish2016confidence,dwork2018decoupled}.
In terms of the time span within which fairness is considered, other than the analysis merely with respect to a snapshot of reality, the literature also includes fairness analyses in dynamic settings \citep{liu2018delayed,hashimoto2018fairness,heidari2019long,zhang2019group,d2020fairness,zhang2020fair,wen2021algorithms,heidari2021allocating,tang2023tier}.

There are explications on available choices to quantify discrimination and enforce fairness in recent survey papers \citep{romei2014multidisciplinary,loftus2018causal,corbett2018measure,mitchell2018prediction,narayanan2018translation,verma2018fairness,caton2020fairness,chouldechova2020snapshot,makhlouf2020survey,makhlouf2021applicability,mehrabi2021survey,zhang2021fairness,pessach2022review} as well as an investigation into public attitudes towards different notions \citep{saxena2019fairness}.
However, the philosophical and methodological contents of the underlying fairness considerations are often not clearly articulated.
In this article, beyond the aforementioned canonical ways of categorizing fairness notions, we review and reflect on previous characterizations of algorithmic fairness in both static and dynamic settings.
In particular, we consider fairness inquiries with different semantic emphases and present a corresponding flowchart to navigate through various fairness spectra.
We believe disentanglement of discriminations based on the intended fairness semantics is vital toward a precise and reasonable quantification of different types of discrimination, so that we can consider the suitable fairness spectrum to better accomplish the goal.
With an extensive discussion into nuances between different intrinsic goals to achieve, we provide a clear picture to make sure that there is no mismatch between the mission (precisely which type of discrimination we really hope to deal with) and the means (which spectrum of fairness we should consider).

The article proceeds as follows:
in Section \ref{sec:philosophy} we take a quick look into how fairness and justice are approached in philosophical discussions;
in Section \ref{sec:preliminaries} we introduce the notation conventions and provide a brief introduction to causal reasoning;
in Sections \ref{sec:fairness_notions} and \ref{sec:dynamic_fairness} we review commonly used algorithmic fairness notions and present fair machine learning studies in the dynamic setting;
in Section \ref{sec:fairness_spectra} we present different spectra of algorithmic fairness inquiries;
in Section \ref{sec:subtlety_causality} we clarify the role of causality in fairness analysis;
in Section \ref{sec:achieve_fairness} we propose an algorithmic fairness flowchart, from which we can see a clearer picture regarding how we should approach the pursuit of different types of fairness;
we summarize with concluding remarks and future works in Section \ref{sec:conclusion}.

\section{What We Talk About When We Talk About Fairness}\label{sec:philosophy}
As we have seen in Section \ref{sec:introduction}, machine learning literature has proposed a deluge of algorithmic fairness definitions, each of which comes with explicit or implicit assumptions on the discrimination of interest and the corresponding mathematical formulation that captures it.
Justice, as a very closely related topic under a different name than ``(algorithmic) fairness'', has been of significant interest to moral and political philosophers.\footnote{
    It has been recognized that ``justice'' and ``fairness'' are not the same thing (see, e.g., \citet{goldman2015justice}).
    Therefore, instead of using ``justice'' and ``fairness'' interchangeably, throughout this section we follow the terminology used by the referenced work to avoid conceptual misunderstandings.
}
It is therefore not surprising to see fairness notions proposed from the machine learning community echo certain justice considerations in ethical theories.
Several recent works have pointed out the necessity of reflecting on such connections \citep{danks2017algorithmic,binns2018fairness,corbett2018measure,glymour2019measuring,heidari2019moral,abebe2020roles,fazelpour2020algorithmic,barocas2020hidden,kasirzadeh2021use,kong2022intersectionally}.

In this section, we first present an empirical scenario as a motivating example.
Then, by listing various fairness-related questions that one might be interested in asking, we lay out different aspects of justice that are formalized and considered in moral and political philosophy.
Here we do not intend to give an overview of theories of justice (see, e.g., \citet{sep-justice}).
Instead, we would like to humbly borrow the wisdom of the rich literature of theories of justice to present a big picture regarding what we are talking about when we talk about algorithmic fairness.

In particular, we examine conceptual dimensions, scopes, and overarching theoretical frameworks.
We would like to demonstrate how one can benefit from a rather rich literature of theories of justice and reflect on the current literature of algorithmic fairness.
The demonstration serves as a starting point, from which one can think about intuitions, (implicit) assumptions, and expectations involved in technical treatments in a principled way.
The example also serves as a preamble to our detailed discussions on spectra of algorithmic fairness inquires in Section \ref{sec:fairness_spectra}, on subtleties of utilization of causality in fairness analysis in Section \ref{sec:subtlety_causality}, and on achieving algorithmic fairness of different spectra in Section \ref{sec:achieve_fairness}.
\reBlue{}{
    We use ``spectrum'' to indicate different categories fairness considerations.
    We reserve ``perspective'' for the discussion on static and dynamic perspectives of fairness analysis.
}{0em}

\subsection{Music School Admission: A Motivating Example}\label{sec:motivating_example}
Let us consider a music school admission example.
Each year, the music school committee considers the admission of applicants to the violin performance program based on their personal information, educational background, instrumental performance, and so on.
The committee also has access to the transcripts of previously admitted students together with their aforementioned information when they applied to the program.

When talking about ``fairness'' in this empirical scenario, based on individuals' intuitive understanding or expectation of fairness, different people may ask different questions, as shown in a non-exhaustive list below:
\begin{enumerate}[label=Question \arabic*, align=left]
    \item (\textit{Ideal} or \textit{Nonideal} methodologies) \label{q:ideal_nonideal}
        When we evaluate fairness of the admission, do we need to first construct an \textit{ideal} world where the admission is fair and to which we then compare our current reality?
        Or do we cope with injustices in the current world and try to move to something better, e.g., a less biased admission in the future?
    \item (\textit{Corrective} or \textit{Distributive} objectives) \label{q:corrective_distributive}
        Are we discussing fairness of the admission for the purpose of \textit{correcting} potentially discriminatory historical decisions, e.g., by admitting a student that was wrongfully denied previously?
        Or are we focusing on \textit{distributing} admission opportunities among current applicants?
    \item (\textit{Procedural} or \textit{Substantive} emphases) \label{q:procedual_substantive}
        Are we considering fairness in terms of how the committee produces the admission decisions, i.e., the decision-making \textit{procedure}? Or do we only care about what the final decision outcomes look like, i.e., who are admitted to the music school this year?
    \item (\textit{Comparative} or \textit{Non-Comparative} considerations) \label{q:comparative_noncomparative}
        Does the fairness consideration involve \textit{comparisons} among individuals, e.g., to compare the decisions received by two applicants who appear to be roughly equally qualified?
        \textcolor{RevisionText}{
            Or are we considering applications separately, i.e., the decision received by one applicant does not affect other applications?
        }
        \reOrange{}{
            In light of the suggestion, we have added a quick example of the \textit{non-comparative} question.
        }{0em}
    \item (The scope of fairness inquiries) \label{q:scope}
        Is the fairness consideration limited to the relationship between the music school and applicants?
        Or are we concerned with a broader \textit{scope} on which the admission decision might have an influence, e.g., the future development of students and their impact on the entire community?
\end{enumerate}

In the rest of this section, we will use this example to demonstrate the connections between intuitive understandings of fairness and discussions of justice in ethical theories.
We will revisit this running example in Section \ref{sec:fairness_spectra}, where we provide additional inquires from a technical treatment point of view and reflect on different spectra of algorithmic fairness inquiries.

\subsection{Conceptual Dimensions, Scopes, and Overarching Theories}\label{sec:example_closer_look}
The idea of justice remains a spotlight of attention in moral, legal, and political philosophy.
As we have seen in the motivating example presented in Section \ref{sec:motivating_example}, there are various fairness inquiries one might be interested in conducting, each of which reveals specific aspects of fairness or justice one would like to pursue.
It is therefore desirable to look at a big picture of ways in which fairness or justice has been approached in ethical theories, so that our discussion can be principled before diving into technical treatments (which will be discussed in the later part of our article).
Following \citet{sep-justice}, we examine conceptual dimensions (Section \ref{sec:conceptual_dimension}), scopes (Section \ref{sec:scope_fairness}), and overarching theoretical frameworks (Section \ref{sec:overarching_framework}) of theories of justice.

\subsubsection{\textbf{The Conceptual Dimensions of Justice}}\label{sec:conceptual_dimension}
~ \\
In this subsection, let us take a look at four essential contrasts in the conceptual apprehension of fairness or justice \citep{sep-justice}, in particular, the \textit{ideal} and \textit{nonideal} methodologies (e.g., \ref{q:ideal_nonideal}), the \textit{corrective} and \textit{distributive} objectives (e.g., \ref{q:corrective_distributive}), the \textit{procedural} and \textit{substantive} emphases (e.g., \ref{q:procedual_substantive}), and the \textit{comparative} and \textit{non-comparative} considerations (e.g., \ref{q:comparative_noncomparative}).

\paragraph{\textit{Ideal} and \textit{Nonideal} Methodologies}
There are two methodological approaches in political philosophy.
The \textit{ideal} approach advances ideal principles according to which a perfectly just (ideal) world operates.
For example, the ``difference principle'', a principle proposed by \citet{rawls1971theory, rawls2001justice} that requires social and economical inequalities to be regulated so that they work to the greatest benefit of the least advantaged member of the society, counts as an \textit{ideal} principle of justice.
The \textit{nonideal} approach, on the other hand, does not posit principles and ideals for a perfectly just society.
Instead, one needs to cope with injustices in the current world and try to move to something better.
For example, as proposed by \citet{anderson2010imperative}, one can evaluate the mechanisms that cause the problem of injustice, as well as responsibilities of different agents to alter these mechanisms, to determine what ought to be done and who should be charged.
Recently, \citet{fazelpour2020algorithmic} have discussed the connection between fair machine learning and the literature on \textit{ideal} and \textit{nonideal} methodological approaches in political philosophy.

\paragraph{\textit{Corrective} and \textit{Distributive} Objectives}
In terms of the objective of fairness inquiries, the contrast between \textit{corrective} and \textit{distributive} justice can date back to Aristotle (\textit{The Nicomachean Ethics, Book V}).
The \textit{corrective} objective of justice concerns a bilateral relationship between the wrongdoer and its victim, emphasizing the remedy that restores the victim to the status before the wrongful behavior occurred.
In contrast, the \textit{distributive} objective of justice involves a multilateral relationship, and formulates justice as a principle to distribute goods of various kinds to individuals.
While \textit{corrective} justice appears more frequently in law practices, current algorithmic fairness literature largely focuses on \textit{distributive} objectives of justice, e.g., the distribution of admission opportunities in our music school example.

\paragraph{\textit{Procedural} and \textit{Substantive} Emphases}
The contrast between the \textit{procedural} and \textit{substantive} emphases reflects different determinants of justice, namely, the justice defined in terms of the procedure itself (e.g., the process how admission committee make the decision) and the justice defined on the substantive outcome (e.g., the final admission decisions of the music school committee).
The distinction between \textit{Disparate Impact} (with a \textit{substantive} emphasis) and \textit{Disparate Treatment} (with a \textit{procedural} emphasis) has been established in law (e.g., Title VII of the 1964 Civil Rights Act) and discussed in the era of big data \citep{barocas2016big,watkins2022four}.
Thanks to the development of causal analysis \citep{spirtes1993causation,pearl2009causality,peters2017elements,hernan2020causal}, fairness in machine learning literature has witnessed extended and ongoing efforts on mathematically formulating and empirically regulating discriminations, both \textit{procedural} and \textit{substantive} ones, which we will see in more detail in Section \ref{sec:preliminaries}.

\paragraph{\textit{Comparative} and \textit{Non-Comparative} Considerations}
Justice can take \textit{comparative} and \textit{non-comparative} forms of considerations.
\textit{Comparative} justice requires one to examine what others can claim when determining what is due to an individual, while \textit{non-comparative} justice determines what is due to an individual merely based on his/her relevant qualities.
In our music school admission example, a fairness inquiry of \textit{comparative} consideration may examine how the admission decision received by one applicant or one demographic group, compared to those received by other applicants or demographic groups.
An inquiry of \textit{non-comparative} consideration may concern whether the decision received by an individual truly respect his/her ability to succeed in the violin program.

\subsubsection{\textbf{The Scope of Justice}}\label{sec:scope_fairness}
~ \\
In Section \ref{sec:conceptual_dimension} we have seen contrasts in conceptual apprehensions of justice.
An important parallel question to ask is when, and to whom, we should apply the concepts or principles of justice.

\paragraph{\textit{Local} and \textit{Global} Views}\label{sec:local_global}
A \textit{local} view argues that principles of justice apply only among individuals who stand in a certain relationship to each other and that the scope is limited to those within such a relationship, e.g., relational theory of justice \citep{rawls1971theory} and local justice \citep{elster1992local,nagel2005problem}.
In our running example, the discussion of fairness limited within the scope of music school admission itself is a \textit{local} view of algorithmic fairness, where the relationship only involves the music school and its applicants.
However, one can consider a broader scope of fairness and ask, for example, what is the long-term impact of current admission decision on potential future developments of applicants as well as their contributions to the society.

\subsubsection{\textbf{The Overarching Theoretical Frameworks to Discuss Justice}}\label{sec:overarching_framework}
~ \\
In this subsection, we present three theoretical frameworks in terms of which justice can be understood, namely, \textit{Utilitarianism}, \textit{Contractarianism}, and \textit{Egalitarianism}.
We note that we highlight these frameworks since they are the more influential ones that have been implicated one way or another in current algorithmic fairness literature.
These frameworks are not intended to be mutually exclusive or exhaustive.

\paragraph{\textit{Utilitarian} Perspective of Justice}
On a high level, \textit{Utilitarianism} aims to maximize the overall welfare, and to bring about the greatest amount of good in terms of the aggregated utilities.
It has been recognized that pure \textit{utilitarianism} is not the final answer to fairness because of several obstacles it faces \citep{rawls1971theory,rawls2001justice,dworkin2002sovereign}:
the ``currency'' of justice or fairness should take the form of benefits/burdens, i.e., the means to gain happiness rather than happiness/unhappiness itself as in \textit{Utilitarianism};
\textit{Utilitarianism} evaluates outcome in terms of the aggregated overall utilities, instead of how utilities are distributed among individuals;
the evaluation is only with respect to the consequences without any consideration about how the consequence is derived in the first place.

\paragraph{\textit{Contractarian} Perspective of Justice}
\textit{Contractarian} philosophers approach justice by looking for (hypothetical) principles in forms of agreements that institutions and individuals all commit to.
David Gauthier characterizes the social contract as a bargain between rational agents and presents the principle of \textit{Minimax Relative Concession} \citep{gauthier1987morals};
John Rawls presents the scenario where people know that their ``conceptions of the good'' are in general different, but at the same time, each individual's conception of the good is placed behind ``a veil of ignorance'' \citep{rawls1971theory,rawls2001justice};
T. M. Scanlon aims to account for ``what we owe to each other'' and presents the idea of justice as a general agreement where no individual, that is informed and unforced, could reasonably reject \citep{scanlon2000we}.

\paragraph{\textit{Egalitarian} Perspective of Justice}
On a high level, \textit{Egalitarianism} aims to establish some sorts of equality.
To a certain extent, equality could act as a default when we intuitively comprehend the idea of fairness and justice.
A natural question faced by \textit{Egalitarianism} is how to make the idea of fairness as equality more specific and reasonable in different contexts.
\textit{Responsibility-sensitive Egalitarianism} approaches this question by treating equal distribution (of opportunities) as a starting point, and allowing for departures from the equality baseline if such departures result from responsible choices of individuals \citep{mason2006levelling,knight2011responsibility};
\textit{Luck Egalitarianism}, as one type of \textit{Responsibility-sensitive Egalitarianism}, adds an additional restriction that the inequalities resulting from brute luck should be constrained \citep{arneson1989equality};
the debates over the role played by \textit{luck} and \textit{desert} also remain a major strand in \textit{Egalitarianism} considerations \citep{anderson1999point,miller2001principles,cohen2009fairness}.

\subsection{Remark: Theories of Justice and Notions of Algorithmic Fairness}
In Section \ref{sec:example_closer_look} we have seen how theories of justice can shed light on various aspects one might consider when discussing ``fairness'' in our running example of music school admission (Section \ref{sec:motivating_example}).
As we shall see in Section \ref{sec:fairness_notions} where we review a non-exhaustive list of definitions of fairness in machine learning literature, ideas of justice often echo in the intuitions behind the proposed algorithmic fairness notions.

\section{Technical Preliminaries}\label{sec:preliminaries}
In this section, we first present the notation conventions used throughout the article in Section \ref{sec:notation}.
Then we present a brief introduction to causal reasoning in Section \ref{sec:causal_modeling}.

\subsection{Notations}\label{sec:notation}
We use uppercase letters to refer to variables, lowercase letters to refer to specific values that variables can take, and calligraphic letters to refer to domains of value.
For instance, we denote the protected feature by $A$ with domain of value $\Acal$, additional (observable) feature(s) by $X$, with domain of value $\Xcal$, ground truth (label) variable by $Y$ and its predictor by $\Yhat$, with domain of value $\Ycal$.

Throughout the article, without loss of generality we assume that there is only one protected feature and one ground truth variable for the purpose of simplifying notation.
Since the protected feature (e.g., race, sex, ratio of ethnic groups within community) and the ground truth variable (e.g., recidivism, annual income) can be discrete or continuous, we do not assume discreteness of the corresponding variables.

There might be additional technical considerations for certain fairness notions to be able to apply in different practical scenarios, for instance,
    the phenomenon of \textit{fairness gerrymandering} \citep{kearns2018preventing,kearns2019empirical} and the quantification of \textit{differential fairness} \citep{foulds2020intersectional} when considering subgroups formed by structured combinations of protected features (the theory of ``intersectionality'' \citep{crenshaw1990mapping,bright2016causally}),
    the challenge introduced by unobserved protected features during learning \citep{hashimoto2018fairness,chen2019fairness} and auditing \citep{awasthi2021evaluating},
    and the risks and opportunities involved in the data collection of demographic information \citep{andrus2021we,andrus2022demographic}.
However, these challenges will not impede us from discussing and reflecting on the intuitions and insights behind fairness notions.

\subsection{Causal Modeling}\label{sec:causal_modeling}
Since we will review commonly used causal notions of fairness and discuss subtleties regarding the role played by causality in fairness analysis, we give a brief introduction to causal modeling and inference in this section.\footnote{
Another field in the causality study is causal discovery where the primary goal is to recover the causal relations among variables from the data \citep{spirtes1993causation,spirtes1995causal,chickering2002optimal,shimizu2006linear,zhang2009identifiability,zhang2011kernel,zhang2017causaldiscovery}.
Causal discovery is not directly related to characterization of fairness in machine learning and therefore is not reviewed in this article.
}
Readers that are already familiar with the related topics may feel free to skip the content.

\subsubsection{\textbf{Definition and Representation of Causality}}
~ \\
For two random variables $X$ and $Y$, we say that $X$ is a \textit{direct cause} of $Y$ if there is a change of distribution for $Y$ when we apply different \textit{interventions} on $X$ while holding all other variables fixed \citep{spirtes1993causation,pearl2009causality}.
We can represent a causal model with a tuple $(U, V, \Fbf)$ such that:
\begin{enumerate}[label=(\arabic*)]
    \item $V$ is a set of observed variables involved in the system of interest;
    \item $U$ is a set of exogenous variables that we cannot directly observe but contains the background information representing all other causes of $V$ and jointly follows a distribution $P(U)$;
    \item $\Fbf$ is a set of functions, also known as structural equations, $\{ f_1, f_2, \ldots, f_n\}$ where each $f_i$ corresponds to one variable $V_i \in V$ and is a mapping $U \cup V \setminus \{V_i\} \rightarrow V_i$.
\end{enumerate}
The triplet $(U, V, \Fbf)$ is known as the \textbf{structural causal model (SCM)}.
We can also capture causal relations among variables via a \textbf{directed acyclic graph (DAG)} $\Gcal$, where nodes (vertices) represent variables and edges represent functional relations between variables and the corresponding direct causes, i.e., observed parents and unobserved exogenous terms.

\textcolor{RevisionText}{  
Here, by representing causal relations via a DAG, we explicitly limit the consideration within the scope of acyclic SCMs (also known as recursive SCMs).\footnote{
    The causal graphs discussed in this article are limited to DAGs, and causal models represented by cyclic graphs are beyond the scope of this article.
}
There are several reasons behind this modeling choice.
To begin with, to the best of our knowledge, current algorithmic fairness literature only considers acyclic SCMs for both static and dynamic settings, and the proposed definitions are largely with respect to causal modeling based on DAGs.
Therefore, when presenting the current literature, we intend to align with this default modeling choice to avoid potential misunderstandings.
Furthermore, limiting the scope of discussion within acyclic SCMs involves additional technical considerations.
The class of acyclic SCMs is a well-studied subclass of SCMs.
Although acyclic SCMs do not have the capacity to model systems with causal cycles, e.g., equilibrium status of dynamic processes, acyclic SCMs have convenient properties including, but not limited to, the uniqueness of the induced distribution, the closedness under (perfect) interventions, the closedness under marginalizations which respect latent projection, and the obedience of various Markov conditions \citep{lauritzen1990independence,spirtes1993causation,lauritzen1996graphical,richardson2003markov,evans2016graphs}.
In general, these properties do not hold true in cyclic SCMs \citep{spirtes1995directed,bongers2021foundations}.
}  

\subsubsection{\textbf{Interventions and Counterfactuals}}
~ \\
Following \citet{pearl2009causality}, we use the $do(\cdot)$ operator to denote an intervention, which is a manipulation of the model such that the value of a variable (or a set of variables) is set to specific values regardless of the corresponding structural equation(s), while leaving other structural equations invariant.
For example, the distribution of $Y$ under the intervention $do(X = x)$ where $X \subseteq V$, is denoted by $P \big(Y \mid do(X = x) \big)$, which reads ``the distribution of $Y$ if we were to force $X = x$ in the population (regardless of the value $X$ takes originally).''

The aforementioned intervention can also be carried out through a specific path (or a set of paths), where a path consists of nodes (variables) connected with a directed edge or a flow of directed edges.
For example, let a path $\pi$ from $X$ to $Y$ be a direct path $X \rightarrow Y$ (or an indirect path $X \rightarrow \cdots \rightarrow Y$), then the distribution of $Y$ under the path-specific intervention $do(X = x |_{\pi})$ along the path $\pi$, is denoted by $P \big(Y \mid do(X = x |_{\pi}) \big)$, which reads ``the distribution of $Y$ if we were to force $X = x$ only along the path $\pi$ (the value change of $X$ is transmitted only along that path) and leave the value of $X$ unchanged along other paths that are not $\pi$.''\footnote{
    With a slight abuse of the notation, if there is no confusion, then we may also use $\pi$ to denote a set of paths of interest.
    The path-specific intervention with respect to the set of paths involves transmitting the value change of $X$ along all paths in the set.
}
\reBlue{}{We add footnote 4 to define $do(\cdot)$ with respect to a set of paths.}{0em}

The full knowledge about the structural equations $\Fbf$ is a rather strong assumption, but it also allows us to infer counterfactual quantities.
For example, let $O, X \subseteq V$ with an observation $O = o$, the counterfactual distribution of $Y$ if $X$ had taken value $x$ is denoted by $P \big( Y_{X \leftarrow x} (U) \mid O = o \big)$, which reads ``the distribution of $Y$ had $X$ been set to $x$ given that we actually observe $O = o$.''
The inference of the counterfactual quantity $P \big( Y_{X \leftarrow x} (U) \mid O = o \big)$ involves a three-step procedure (as explained in more detail in \citet{pearl2009causality}):
\begin{enumerate}[label=(\arabic*)]
    \item \textbf{Abduction}: for a given prior on $U$, compute the posterior distribution of $U$ given the observation $O = o$;
    \item \textbf{Action}: substitute the structural equation that determines the value of $X$ with the intervention $X = x$ and get modified set of structural equations $\Fbf_{\mathrm{modify}}$;
    \item \textbf{Prediction}: compute the distribution of $Y$ using $\Fbf_{\mathrm{modify}}$ and the posterior $P(U \mid O = o)$.
\end{enumerate}

The counterfactual quantities can also be defined in a path-specific manner.
For example, suppose that the intervention on $X$ is only transmitted through the path $\pi$, then the path-specific counterfactual distribution of $Y$ if $X$ had taken value $x$ only along the path $\pi$ is denoted by $P \big( Y_{X \leftarrow x | \pi} (U) \mid O = o \big)$, which reads ``the distribution of $Y$ had $X$ been set to $x$ only along the specific path $\pi$ given that we actually observed $O = o$.''

The identifiability of various causal quantities has been extensively studied in the literature \citep{tian2002general,avin2005identifiability,huang2006identifiability,shpitser2006identification,shpitser2007counterfactuals,shpitser2008complete}.

\section{Instantaneous Notions of Algorithmic Fairness}\label{sec:fairness_notions}
In Section \ref{sec:philosophy}, we present normative considerations of fairness and justice, in this section, we present technical details of a non-exhaustive list of instantaneous fairness notions proposed in the literature.
Here by ``instantaneous'' we are referring to the fact that the fairness inquiry is with respect to a given snapshot of reality.
This characteristic is also called ``static'' fairness in the literature \citep{d2020fairness}.
Considering the fact that instantaneous fairness notions are also considered in dynamic settings in the literature, to avoid confusion, we use the term ``instantaneous'' to indicate that the fairness notion is not explicitly time-dependent, and we reserve the term ``static'' to distinguish from ``dynamic'' when discussing different settings where fairness inquiries take place (we review fairness considerations in the dynamic setting in Section \ref{sec:dynamic_fairness}).
When presenting various previously proposed fairness notions, we unify the notations for consistency while keeping their meanings intact.

\subsection{Demographic Parity}\label{sec:notion_DP}
\textit{Demographic Parity}, also known as \textit{Statistical Parity}, is one of the earliest fairness notions proposed in the literature \citep{calders2009building,dwork2012fairness,zemel2013learning,feldman2015certifying}.
In the context of binary classification ($\Ycal = \{0, 1\}$), \textit{Demographic Parity} requires that the ratio of positive decisions among different groups is equal:
\begin{equation}
    \forall a, a' \in \Acal: ~
    P(\Yhat = 1 \mid A = a) = P(\Yhat = 1 \mid A = a').
\end{equation}
In general contexts, \textit{Demographic Parity} is characterized via the independence between the prediction $\Yhat$ and the protected feature $A$.

\begin{definition}[Demographic Parity]\label{def:DP}
    We say that a predictor $\Yhat$ is fair in terms of \textit{Demographic Parity} with respect to the protected feature $A$, if $\Yhat$ is independent from $A$, i.e., $\Yhat \indep A$.
\end{definition}
While it is intuitive to characterize fairness through the aforementioned independence, the notion has significant drawbacks \citep{dwork2012fairness}.
For instance, when there is unobjectionable dependence between the ground truth $Y$ and the protected feature $A$, i.e., $Y \nindep A$, by definition the perfect predictor is also dependent on $A$ ($\Yhat \nindep A$ since $\Yhat = Y$).
It is not intuitive why we should rule out the perfect predictor (although this might not be achievable in reality) for the sake of satisfying the \textit{Demographic Parity} fairness requirement on the prediction even if we allow $Y \nindep A$ in the data.

\subsection{Equalized Odds}\label{sec:notion_EOdds}
In light of the limitation of \textit{Demographic Parity}, \citet{hardt2016equality} propose the \textit{Equalized Odds} notion of fairness.
In the context of binary classification, \textit{Equalized Odds} requires that the \textbf{True Positive Rate (TPR)} and \textbf{False Positive Rate (FPR)} of each group match the population TPR and FPR respectively:
\begin{equation}
    \forall a \in \Acal, y \in \{0, 1\}: ~
    P(\Yhat = 1 \mid A = a, Y = y) = P(\Yhat = 1 \mid Y = y).
\end{equation}
In general contexts, this notion is characterized by stating the conditional independence between the prediction $\Yhat$ and the protected feature $A$ given the ground truth of the target $Y$.

\begin{definition}[Equalized Odds]\label{def:equalized_odds}
    We say that a predictor $\Yhat$ is fair in terms of \textit{Equalized Odds} with respect to the protected feature $A$ and the outcome $Y$, if $\Yhat$ is conditionally independent from $A$ given $Y$, i.e., $\Yhat \indep A \mid Y$.
\end{definition}
The intuition behind this group-level fairness notion is that, once we know the true value of the target (in the hypothetical ideal world), the additional information of the value of the protected feature should not further alter our prediction results.

\subsection{Predictive Parity}\label{sec:notion_PP}
First proposed by \citet{dieterich2016compas}, \textit{Predictive Parity} is another group-level fairness notion, which is also referred to as calibration, \textit{Test Fairness} \citep{chouldechova2017fair} and \textit{No Disparate Mistreatment} \citep{zafar2017fairness}.
In the context of binary classification, \textit{Predictive Parity} requires that among those whose predicted value is positive (negative), their probability of actually having a positive (negative) label should be the same regardless of the value of the protected feature:
\begin{equation}
    \forall a \in \Acal, \yhat \in \{0, 1\}: ~
    P(Y = 1 \mid A = a, \Yhat = \yhat) = P(Y = 1 \mid \Yhat = \yhat).
\end{equation}
Similar to \textit{Equalized Odds}, \textit{Predictive Parity} can also be characterized through the conditional independence relation among $(A, Y, \Yhat)$.
\reBlue{}{
    Compared to Equations (1, 2, 3), in Equations (4, 5, 6, 7, 8, 10, 11, 13), we use the form of P1 - P2 = 0 instead of P1 = P2 since we would like to align with the requirement of fairness notions that causal effects are set to zero.
}{0em}

\begin{definition}[Predictive Parity]\label{def:PP}
    We say that a predictor $\Yhat$ is fair in terms of \textit{Predictive Parity} with respect to the protected feature $A$ and the outcome $Y$, if $Y$ is conditionally independent from $A$ given $\Yhat$, i.e., $Y \indep A \mid \Yhat$.
\end{definition}

Although they look similar, \textit{Demographic Parity}, \textit{Predictive Parity} and \textit{Equalized Odds} are actually incompatible with each other.
It is shown independently by \citet{kleinberg2017inherent} and \citet{chouldechova2017fair} that any two out of three conditions can not be attained at the same time except in very special cases.
For example, one can achieve \textit{Demographic Parity} and \textit{Equalized Odds} at the same time when $A$ and $Y$ are independent, or $\Yhat$ is a trivial predictor (always constant or completely random).

In fact, the aforementioned incompatibility results also provide additional insights regarding the widely observed trade-offs between fairness and utility \citep{kamiran2012data,romei2014multidisciplinary,feldman2015certifying,chouldechova2017fair,berk2017convex,corbett2017algorithmic,kleinberg2017inherent,menon2018cost,agarwal2018reductions,mary2019fairness,wick2019unlocking,baharlouei2020renyi}.
According to the information bottleneck principle \citep{tishby2000information,tishby2015deep}, we would like the predicted outcome $\Yhat$ to be an information bottleneck through which we capture as much information as possible between the target variable $Y$ and features including, but not limited to, the protected feature.
The information bottleneck principle aligns with the conditional independence relation required by \textit{Predictive Parity} notion of fairness (Definition \ref{def:PP}).
This indicates that the unconstrained optimization can have \textit{Predictive Parity} fairness as a byproduct.
In other words, there is no conflict in principle, and therefore, no trade-off, between \textit{Predictive Parity} fairness and unconstrained optimizations.
This phenomenon is also referred to as an ``implicit fairness criterion of unconstrained learning'' \citep{liu2019implicit}.
This also indicates that notions that are incompatible with \textit{Predictive Parity} will necessarily involve trade-offs between fairness and accuracy compared to unconstrained optimizations.

\subsection{Associative Notions of Individual-level Fairness}\label{sec:notion_IF}
\color{RevisionText}  
Apart from associative notions of group-level fairness (e.g., \textit{Demographic Parity}, \textit{Equalized Odds}, \textit{Predictive Parity}), individual-level fairness notions that are based on associative relations among variables are also proposed in the literature.\footnote{
    We will see causal notions of individual-level fairness in Section \ref{sec:notion_CF}.
}
A canonical example of the individual-level associative notion is \textit{Individual Fairness} proposed by \citet{dwork2012fairness}.
The intuition behind \textit{Individual Fairness} is that we want similar predicted outcome for similar individuals according to specific similarity metrics.
Apart from various choices of the metric, there are different formulations in the literature to mathematically capture the aforementioned intuition.
For completeness we present both the \textit{Lipschitz Mapping}-based formulation \citep{dwork2012fairness} and the $(\epsilon-\delta)$ language-based formulation \citep{georgejohn2020verifying} of \textit{Individual Fairness}.

\begin{definition}[(Lipschitz Mapping) Individual Fairness]\label{def:IF_mapping}
    We say that a mapping $h: \Acal \times \Xcal \rightarrow \Ycal$ satisfies \textit{Individual Fairness} if it is $L$-Lipschitz with respect to appropriate metrics on the domain, i.e., $\Acal \times \Xcal$, and the codomain, i.e., $\Ycal$:
    \begin{equation*}
        \forall (a, x), (a', x') \in \Acal \times \Xcal: ~
        d_{\Ycal} \big( h(a, x), h(a', x') \big)
        \leq L \cdot d_{\Acal \times \Xcal} \big( (a, x), (a', x') \big).
    \end{equation*}
\end{definition}

\begin{definition}[($\epsilon - \delta$) Individual Fairness]\label{def:IF_language}
    Let us consider $\epsilon \geq 0$, $\delta \geq 0$, and a mapping $h: \Acal \times \Xcal \rightarrow \Ycal$.
    We say that $h$ satisfies \textit{Individual Fairness} if
    \begin{equation*}
        \forall (a, x), (a', x') \in \Acal \times \Xcal: ~
        d_{\Acal \times \Xcal} \big( (a, x), (a', x') \big) \leq \epsilon
        \implies d_{\Ycal} \big( h(a, x), h(a', x') \big) \leq \delta.
    \end{equation*}
\end{definition}

While \textit{Individual Fairness} is general enough to be applicable in various practical scenarios, the specification of the similarity metric is not often straightforward.
Recent literature has explored ways to achieve individual fairness of different flavors \citep{friedler2016possibility,joseph2016fairness,kearns2017meritocratic,gillen2018online,heidari2018fairness,kim2018fairness,sharifi2019average,ilvento2020metric,mukherjee2020two,ruoss2020learning,yurochkin2020training,jung2021algorithmic,yurochkin2021sensei,benussi2022individual}.
The connection between group-level and individual-level fairness notions beyond their seemingly apparent conflicts also draws attentions \citep{speicher2018unified,binns2020apparent,fleisher2021what}.
\color{black}  

\subsection{No Direct/Indirect Discrimination}\label{sec:notion_no_direct_indirect}
Fairness notions presented in Sections \ref{sec:notion_DP} - \ref{sec:notion_IF} are based on associative relations among variables.
Going beyond these observational criteria, it is desirable if we can further capture the structure of the data generating process by making use of causal modeling.

In the legislation literature, discrimination is commonly divided into two categories: direct discrimination (e.g., rejecting a well-qualified loan applicant only because of the demographic identity) and indirect discrimination (e.g., refusing service to areas with certain Zip code).
The motivation behind detecting indirect discrimination is that: among the non-protected attributes $X$, there is a set of attributes whose usage may still remain (potentially) unjustified, i.e., redlining attributes $R$, although they are not the protected feature itself.
In the language of causal reasoning, given a causal graph, we can start from the node for the protected feature and trace along the paths all the way to the node of interest by following the arrowheads in the graph.
Therefore, we can characterize direct and indirect discrimination as different path-specific causal effects with respect to the protected feature \citep{pearl2009causality,zhang2016situation,zhang2017achieving,zhang2017anti,zhang2017causal,nabi2018fair,zhang2018fairness,nabi2019learning,nabi2019optimal}.

\begin{definition}[No Direct Discrimination]\label{def:no_direct_discrimination}
    Let us denote as $\pi_d$ the path set that contains only the direct path from the protected feature $A$ to the predictor $\Yhat$, i.e., $A \rightarrow \Yhat$.
    We say that a predictor $\Yhat$ is fair in terms of \textit{No Direct Discrimination} with respect to the protected feature $A$ and the path set $\pi_d$, if for any $a, a' \in \Acal$ and $\yhat \in \Ycal$ the $\pi_d$-specific causal effect of the change in $A$ from $a$ to $a'$ on $\Yhat = \yhat$ satisfies:
    \begin{equation}
        P \big( \Yhat = \yhat \mid do(A = a'|_{\pi_d}) \big)
        - P \big( \Yhat = \yhat \mid do(A = a) \big) = 0.
    \end{equation}
\end{definition}

\begin{definition}[No Indirect Discrimination]\label{def:no_indirect_discrimination}
    Let us denote as $\pi_i$ the path set that contains all causal paths from the protected feature $A$ to the predictor $\Yhat$ which go though redlining attributes $R$, i.e., each path within the set $\pi_i$ includes at least one node from $R$.
    We say that a predictor $\Yhat$ is fair in terms of \textit{No Indirect Discrimination} with respect to the protected feature $A$ and the path set $\pi_i$, if for any $a, a' \in \Acal$ and $\yhat \in \Ycal$ the $\pi_i$-specific causal effect of the change in $A$ from $a$ to $a'$ on $\Yhat = \yhat$ satisfies:
    \begin{equation}
        P \big( \Yhat = \yhat \mid do(A = a'|_{\pi_i}) \big)
        - P \big( \Yhat = \yhat \mid do(A = a) \big) = 0.
    \end{equation}
\end{definition}

Motivated by the idea of capturing discrimination through different types of causal effects of the protected feature on the predictor, similar notions are also proposed by \citet{kilbertus2017avoiding} to further distinguish different types of attributes that are descendants of the protected feature.
In particular, for attributes that are influenced by the protected feature $A$ in a manner that we deem as non-discriminatory, i.e., resolving variables, the path-specific causal effects of $A$ on $\Yhat$ through these attributes are ``resolved.''
For attributes that are influenced by $A$ in an unjustifiable way, i.e., proxy variables, the path-specific causal effects of $A$ on $\Yhat$ through these attributes are ``unresolved.''

\begin{definition}[No Unresolved Discrimination]\label{def:no_unresolved_discrimination}
    We say that a predictor $\Yhat$ is fair in terms of \textit{No Unresolved Discrimination}, if each path from $A$ to $\Yhat$ is blocked by a resolving variable in the corresponding causal graph.
\end{definition}

\begin{definition}[No Proxy Discrimination]\label{def:no_proxy_discrimination}
    We say that a predictor $\Yhat$ is fair in terms of \textit{No Proxy Discrimination} with respect to a proxy $R$, if for any $r, r' \in \mathcal{R}$ and $\yhat \in \Ycal$:
    \begin{equation}
        P \big( \Yhat = \yhat \mid do(R = r) \big)
        - P \big( \Yhat = \yhat \mid do(R = r') \big) = 0.
    \end{equation}
\end{definition}

Similar to related notions like ``explanatory feature'' \citep{kamiran2013quantifying}, ``redlining attribute'' \citep{zhang2017causal}, and ``admissible variables'' \citep{salimi2019interventional}, the notion of ``resolving variable'' and ``proxy variable'' are just descendants of $A$ with different user-specified characteristics.
Compared to \textit{No Indirect Discrimination}, although \textit{No Proxy Discrimination} is also capturing indirect discrimination through proxy variables, the intervention based on the proxy variable is conceptually easier to parse compared to the intervention on the protected feature itself.
The protected feature, e.g., gender or race, is a deeply rooted personal property and it is impossible to perform a randomized trial \citep{vanderweele2014causal}.

\subsection{Counterfactual Fairness}\label{sec:notion_CF}
So far, the causal notions of fairness (\textit{No Direct/Indirect Discrimination}, \textit{No Unresolved Discrimination}, \textit{No Proxy Discrimination}) are quantifying the discrimination on the group level.
\textit{Counterfactual Fairness} proposed by \citet{kusner2017counterfactual}, compared to previous ones, is more fine-grained since it captures individual-level notion of fairness.
\reOrange{}{
    We follow the suggestion of including a separate section for associative notions of individual fairness.
    The added material can be found in Section \ref{sec:notion_IF}.
}{0em}

In Section \ref{sec:notion_IF}, we have seen the characterization of individual-level fairness by making use of associative relations among variables.
\textit{Counterfactual Fairness}, however, approaches the individual-level fairness problem from a different angle by making use of causal relations among variables.
In particular, the intuition behind \textit{Counterfactual Fairness} is that a decision is fair towards an individual if the decision remains the same in the actual world (the current reality) and a properly defined counterfactual world (the hypothetical world where this individual had a different demographic property).

\begin{definition}[Counterfactual Fairness]\label{def:counterfactual_fairness}
    Given a causal model $(U, V, \Fbf)$ where $V$ consists of all features $V \coloneqq \{A, X\}$, we say that a predictor $\Yhat$ is fair in terms of \textit{Counterfactual Fairness} with respect to the protected feature $A$, if for any $a, a' \in \Acal, x \in \Xcal, \yhat \in \Ycal$ the following holds true:
    \begin{equation}
        P \big( \Yhat_{A \leftarrow a}(U) = \yhat \mid A = a, X = x \big)
        - P \big( \Yhat_{A \leftarrow a'}(U) = \yhat \mid A = a, X = x \big) = 0.
    \end{equation}
\end{definition}

\subsection{Path-specific Counterfactual Fairness}\label{sec:notion_PC_CF}
The \textit{Path-specific Counterfactual Fairness} notion \citep{chiappa2019path,wu2019pc} shares the similar intuition with \textit{Counterfactual Fairness} and captures the difference in decision between the actual world and a counterfactual world.\footnote{
    \citet{wu2019pc} uses the abbreviated term \textit{PC Fairness} to denote a unified formula for various causal notions of fairness.}
Different from \textit{Counterfactual Fairness}, more fine-grained causal effects are utilized by \textit{Path-specific Counterfactual Fairness}---path-specific counterfactual effects, i.e., the counterfactual causal effects are characterized only through unfair paths.

\begin{definition}[Path-specific Counterfactual Fairness]\label{def:pc_fairness}
    Given a causal model $(U, V, \Fbf)$ and a factual observation $O = o$, where $V$ consists of all features $V \coloneqq \{A, X\}$ and $O \subseteq \{A, X, Y\}$, we say that a predictor $\Yhat$ is fair in terms of \textit{Path-specific Counterfactual Fairness (PC Fairness)} with respect to the protected feature $A$ and the path set $\pi$, if for any $a, a' \in \Acal, \yhat \in \Ycal$ the $\pi$-specific counterfactual causal effect of the change in $A$ from $a$ to $a'$ on $\Yhat = \yhat$ satisfies (let $\bar\pi$ denote the set containing all other paths in the graph that are not elements of $\pi$):
    \begin{equation}
        P \big( \Yhat_{A \leftarrow a' | \pi, A \leftarrow a | \bar\pi}(U) = \yhat \mid O = o \big)
        - P \big( \Yhat_{A \leftarrow a}(U) = \yhat \mid O = o \big) = 0.
    \end{equation}
\end{definition}

For different configurations of the observation $O = o$ and the path set of interest $\pi$, \textit{PC Fairness} can capture different types of causal effects, which results in various flavors of fairness notions.
For example, if $\pi$ consists of all paths in the graph and $O = \{A, X\}$, this configuration of \textit{PC Fairness} (for every $O = o$) reduces to \textit{Counterfactual Fairness} \citep{wu2019pc}.

\subsection{Remark: Connect Theories of Justice and Notions of Algorithmic Fairness}
In Section \ref{sec:philosophy} we have seen that ethical arguments about fairness or justice can vary across conceptual dimensions, scopes, and overarching theoretical frameworks.
Although it is less extensively elaborated in the algorithmic fairness literature, the difference in proposed fairness notions reveals the nuances behind different understandings about what and how algorithmic fairness should be captured.

In terms of the overarching theoretical frameworks, on a high level, the commonly used algorithmic fairness notions rest upon specific types of equality, which align with the idea advocated in \textit{Egalitarianism};
at the same time, the practice of performance optimization (with fairness considerations) aligns with \textit{Utilitarian} considerations.

In terms of conceptual dimensions, recent algorithmic fairness notions largely follow the \textit{Ideal} methodology where an ideal principle is advocated regarding what the ideally fair world should look like.
For example, Definition \ref{def:DP} proposes an independence relation as the ideal principle, and Definition \ref{def:counterfactual_fairness} advocates a zero counterfactual causal effect.
The \textit{Nonideal} methodology has attracted attentions in recent algorithmic fairness literature (see, e.g., \citet{fazelpour2020algorithmic}) but is relatively less explored compared to the \textit{Ideal} counterpart.
The distinction between \textit{Procedural} and \textit{Substantive} considerations is well-represented by the distinction between causal and associative notions of algorithmic fairness.
The form of \textit{Comparative} consideration (i.e., to draw comparisons between individuals) echoes in various individual fairness notions (Section \ref{sec:notion_IF}) as well as other notions that are defined on the amount of effort one needs to make in order to get preferable results \citep{heidari2019long,huang2020fairness,vonkugelgen2022fairness}.

In terms of the scope of consideration, the current algorithmic fairness literature primarily focuses on \textit{Local} fairness in the sense that the fairness inquiry is limited to the specific scenario at hand.
In Section \ref{sec:fairness_spectra}, when we present fairness spectra, and in Section \ref{sec:achieve_fairness}, when we present the flowchart for algorithmic fairness, we argue that when fairness inquiries are performed in a closed-loop format, one can potentially further improve fairness to a larger scale, i.e., going beyond \textit{Local} fairness and towards \textit{Global} fairness.

\section{Fairness in Dynamic Settings}\label{sec:dynamic_fairness}
In Section \ref{sec:fairness_notions}, we have seen various fairness notions defined in an instantaneous manner, i.e., with respect to a fixed snapshot of reality.
Considering the ever-present changes happening in practical scenarios, it has been widely recognized that fairness audits in a purely static setting may not serve the purpose of understanding the long-term impact of machine learning algorithms \citep{ensign2017decision,ensign2018runaway,hashimoto2018fairness,heidari2018preventing,hu2018short,kim2018collective,liu2018delayed,bechavod2019equal,elzayn2019fair,heidari2019long,hu2019disparate,kannan2019downstream,milli2019social,mouzannar2019fair,zhang2019group,creager2020causal,d2020fairness,kleinberg2020classifiers,liu2020disparate,zhang2020fair,heidari2021allocating,raab2021unintended,wen2021algorithms,zhang2021fairness,vonkugelgen2022fairness,tang2023tier}.
In this section, we review existing literature on fairness studies in the dynamic setting.

There are application-specific studies in the dynamic fairness literature:
the opportunity allocation in labor market \citep{hu2018short},
a pipeline consisting of college admission followed by hiring \citep{kannan2019downstream},
the opportunity allocation in credit application \citep{liu2018delayed},
and the resource allocation in predictive policing \citep{ensign2018runaway}.

Apart from application-specific studies, the literature has adopted various analyzing frameworks to approach dynamic fairness audits, for instance, the utilization of the P\'olya urn model in incident discovery \citep{hu2018short,ensign2018runaway} and intergenerational mobility analysis \citep{heidari2021allocating},
fairness inquiries conducted through one-step analyses \citep{liu2018delayed,kannan2019downstream,mouzannar2019fair,zhang2019group},
the leverage of \textbf{reinforcement learning (RL)} \citep{sutton2018reinforcement} techniques, e.g., \textbf{multi-armed bandits (MABs)} \citep{joseph2016fairness,joseph2016fair,liu2017calibrated,gillen2018online,li2020hiring,claure2020multi,patil2021achieving,wang2021fairness,tang2021bandit}
and \textbf{Markov decision processes (MDPs)} \citep{jabbari2017fairness,siddique2020learning,zhang2020fair,wen2021algorithms,zimmer2021learning,ge2021towards},
fairness inquiries conducted in online settings (where algorithms improve as new samples arrive sequentially) \citep{heidari2018preventing,bechavod2019equal,elzayn2019fair,bechavod2020metric},
the challenge introduced by domain shifts \citep{schumann2019transfer,singh2021fairness,rezaei2021robust,liu2021induced},
the game-theoretic equilibrium analyses \citep{coate1993will,mouzannar2019fair,liu2020disparate},
the efforts towards providing interpretations of dynamic and long-term fairness via causal modeling \citep{creager2020causal} and simulation studies \citep{d2020fairness}.

The practical application provides a specific context for fairness considerations, depending on which one would expect context-dependent interpretations of technical findings.
This indicates the importance of the modeling choice in the dynamic setting.
In the following subsections, we present common choices of analyzing frameworks, namely, the P\'olya urn model (Section \ref{sec:choice_urn}), the one-step feedback model (Section \ref{sec:choice_one_step}), and the reinforcement learning framework (Section \ref{sec:choice_RL}).
Then in Section \ref{sec:remark_choice_analysis}, we provide a remark on types of dynamics modeled in the literature.

Considering the fact that take-away messages are closely related to modeling choices, when presenting the previously reported results in the literature, we lay out assumptions and modeling choices before summarizing findings and only resort to detailed technical representations when necessary.
Since modeling choices and notations vary across different approaches, in each subsection we follow the original notation scheme used by authors of the referenced work.

\subsection{Choice of Analyzing Framework: P\'olya Urn Model}\label{sec:choice_urn}
In the (generalized) P\'olya urn model, there are two colors of balls, let us say red and black, in the urn.
At each time step, one ball is drawn from the urn, then its color is noted and the ball is replaced.
There is a replacement matrix of the following form:
\begin{equation}\label{equ:urn_matrix}
    \begin{array}{lcc}
        & \text{Red addition} & \text{Black addition} \\
        \text{Sample red} & a & b \\
        \text{Sample black} & c & d \\
    \end{array},
\end{equation}
which governs how urn content is updated.
For example, if the urn follows the replacement dynamics as detailed in Equation (\ref{equ:urn_matrix}), every time we sample a red (black) ball, we replace it and further add $a$ ($c$) more red balls and $b$ ($d$) more black balls into the urn.

\citet{ensign2018runaway} use the P\'olya urn to model the recording and (re-)occurrence of crime incidents in certain neighborhoods.
In particular, they consider an urn that contains two colors of balls (red and black) that correspond to two neighborhoods ($A$ and $B$).
At each time step, the police patrol in neighborhood $A$ ($B$) corresponds to drawing a red (black) ball from the urn, and observing a crime in the neighborhood corresponds to placing a ball of the same color into the urn.
The initially drawn balls will always be replaced before the next time step.
The ratio between counts for red and black balls represents the observed crime statistics, and the long-term distribution of color proportions reflects the modeled long-term belief about crime prevalence in neighborhoods.

A similar instantiation of the P\'olya urn model is also (implicitly) utilized in intergenerational mobility analysis \citep{heidari2021allocating}.
In their model, the population consists of two groups, the advantaged group ($A$) and the disadvantaged group ($D$).
The group identity of an individual is not fixed across the temporal axis.
In each time step, the society can only offer opportunities to an $\alpha$ (fixed) fraction of the population, and the problem at hand is how to allocate this limited amount of opportunities in the society.
Individual with different socioeconomic status (advantaged/disadvantaged) has different probability of succeeding if provided with an opportunity.
Any individual in the disadvantaged group $D$ who succeeds after being offered the opportunity will relocate into the advantaged group $A$.
Then, every individual is replaced with its next generation of the same socioeconomic status, and the aforementioned process continues.
In their standard model, the ``replacement'' of individuals in the new generation is essentially controlled by hyperparameters in the replacement matrix, i.e., the standard P\'olya urn model by setting $a = d = 1$ and $b = c = 0$ in Equation (\ref{equ:urn_matrix}).
If individual's offspring does not perfectly inherit its socioeconomic status, then the generalized P\'olya urn model will be utilized.

\subsection{Choice of Analyzing Framework: One-step Feedback Model}\label{sec:choice_one_step}
Different from analyzing dynamic fairness along multiple time steps, previous works also consider one-step feedback models \citep{liu2018delayed,kannan2019downstream,mouzannar2019fair,zhang2019group}.

\citet{kannan2019downstream} focus on a pipeline consisting of college admission and hiring.
They propose a two-stage model with the hiring result at the end of the pipeline as the single one-step feedback.
\citet{liu2018delayed} utilize a one-step feedback model to study how static fairness notions interact with well-being of agents on the temporal axis.

\citet{mouzannar2019fair} focus on the \textit{Demographic Parity} (Definition \ref{def:DP}) form of affirmative action (fairness intervention) and model at the same time (1) a selection process where the utility-maximizing institution performs binary classification according to the qualification of agents from different groups, and (2) the evolution of group qualifications after imposing the selection with affirmative actions.
In their one-step feedback model, the institution uses a deterministic threshold policy on the one-dimensional summary attribute of the agent at the time step $t$, and this selection process influences the group-level qualification profiles at the time step $t + 1$.

\citet{zhang2019group} focus on the relation between the enforced fairness and group representations, as well as the impact of decision on underlying feature distributions.
They model group representations via a one-step update function, which governs how the expected number of customers in a group at the time step $t + 1$ is determined by quantities at the time step $t$: the expected number of customers from that group, current customer retention rate, and the expected new customers arrivals from that group.

\subsection{Choice of Analyzing Framework: Reinforcement Learning}\label{sec:choice_RL}
Previous works have approached dynamic fairness audits via the framework of MABs.
\citet{joseph2016fairness,joseph2016fair} study dynamic fairness in stochastic and contextual bandits problems.
In their \textit{Meritocratic Fair} definition of fairness, agents of lower qualification are never favored over agents of higher qualification, despite the possible uncertainty of the algorithm.\footnote{The term ``Meritocratic Fairness'' is also utilized as a fairness notion to capture (instantaneous) subgroup fairness \citep{kearns2017meritocratic}, and should not be confused with the dynamic setting considered by \citet{joseph2016fairness,joseph2016fair}.}

\citet{liu2017calibrated} utilize the stochastic MAB framework and adopt the ``treating similar individuals similarly'' \citep{dwork2012fairness} notion of individual fairness.
Here the notion of ``individual'' corresponds to an arm, and two arms are pulled near-indistinguishably if they have a ``similar'' qualification distribution.
\citet{liu2017calibrated} complement the aforementioned work by \citet{joseph2016fairness} by incorporating a smoothness constraint and providing a protection of higher qualifications over lower qualifications in an on-average manner.

\citet{gillen2018online} consider the problem of online learning in linear contextual bandits with an unknown metric-based individual fairness \cite{dwork2012fairness}.
They assume that only weak feedback, one that flags the violation of an unknown similarity metric but without quantification, is available, and provide an algorithm in this adversarial context.

\citet{li2020hiring} view the hiring process as a contextual bandit problem and pay special attention to the balance between ``exploitation'' (selecting from groups with proven hiring records) and ``exploration'' (selecting from under-represented groups to gather information).
\citet{li2020hiring} propose an algorithm that emphasizes exploration by evaluating individuals' statistical upside potential, and highlight the importance of incorporating exploration in decision making in the pursuit of dynamic fairness.

\citet{patil2021achieving} consider the fairness requirement of pulling each arm at least some pre-specified fraction of times in the stochastic MAB problem.
\citet{wang2021fairness} study the fairness of exposure \citep{singh2018fairness} in the online recommending system, and propose a new objective for the stochastic bandits setting to resolve the issue of winner-takes-all allocation of exposure.
\citet{tang2021bandit} consider the setting where past actions can have delayed impacts on arm rewards in the future.
They take into account the runaway feedback issue \citep{ensign2018runaway} due to action history, and study the delayed-impact phenomenon in the stochastic MAB context.

Previous works have also approached dynamic fairness audits via the framework of MDPs.
\citet{jabbari2017fairness} take into consideration the impact of actions on states (environments) and future rewards, and enforce the fairness notion that an algorithm never prefers an action over another if the long-term (discounted) accumulated reward of the latter is higher (\textit{Meritocratic Fair} \citep{joseph2016fairness}).
\citet{siddique2020learning} integrate the \textbf{generalized Gini social welfare function (GGF)} \citep{weymark1981generalized} with \textbf{multi-objective Markov decision processes (MOMDPs)}, where rewards take the form of vector instead of scalar, to impose the specific notion of fairness.
\citet{zimmer2021learning} consider the problem of deriving fair policies in cooperative \textbf{multi-agent reinforcement learning (MARL)}.
\citet{zhang2020fair} consider \textit{Demographic Parity} and \textit{Equal Opportunity} notions of fairness with respect to the dynamics of group-level qualification, in the \textbf{partially observed Markov decision process (POMDP)} setup.
They demonstrate the fact that static fairness notions can result in both improvement and deterioration of fairness depending on the specific characteristics of the POMDP.
\citet{wen2021algorithms} model the feedback effect of decisions as the dynamics of MDPs, and audit fairness with respect to group-conditioned outcomes of agents in terms of \textit{Demographic Parity} and \textit{Equal Opportunity}.
\citet{ge2021towards} consider long-term group-level fairness of exposure \citep{singh2018fairness} with non-fixed group labels in the context of recommending systems, and formulate the recommendation problem as a \textbf{constrained Markov decision process (CMDP)}.


\subsection{Remark: Differences in Modeled Dynamics}\label{sec:remark_choice_analysis}
Apart from common choices of analyzing frameworks presented in Sections \ref{sec:choice_urn}-\ref{sec:choice_RL}, previous dynamic fairness literature also considers different types of user dynamics, for instance,
the retention dynamics of the customer \citep{zhang2019group},
the amplification dynamics of representation disparity \citep{hashimoto2018fairness,ensign2018runaway},
the imitation and replicator dynamics of agents \citep{heidari2019long,raab2021unintended},
the strategic behavior of agents \citep{dong2018strategic,hu2019disparate,milli2019social,kleinberg2020classifiers,estornell2021unfairness},
the algorithmic recourse for agents \citep{ustun2019actionable,joshi2019towards,vonkugelgen2022fairness},
the rational investments of agents \citep{hu2018short,heidari2019long,liu2020disparate},
the intergenerational mobility \citep{heidari2021allocating}.
Considering that a comprehensive literature review of algorithmic fairness inquiries in dynamic settings is beyond the scope of our article, we proceed with reflections on algorithmic fairness in the rest of the article (Sections \ref{sec:fairness_spectra}-\ref{sec:achieve_fairness}).\footnote{
Interested readers please refer to a recent survey on fairness in learning-based sequential decision algorithms \citep{zhang2021fairness}.
}

\section{Different Spectra of Fairness Inquiries}\label{sec:fairness_spectra}
In Sections \ref{sec:fairness_notions} and \ref{sec:dynamic_fairness}, we have surveyed fairness inquiries in both static and dynamic settings.
In this section, we reflect on different spectra of fairness inquiries.
We start by revisiting our running example of music school admission, and focus on the intuition behind each question on the inquiry checklist for fairness in this example.
The reflection is not limited to any particular notion of fairness in the literature.
Instead, we take a step back and think about the exact type of fairness each question is trying to get at by considering, for instance, the unstated assumption, the intended discussing context, and so on.
The categorization of previously proposed notions of fairness, as well as technical details of potential modifications to the notion, will be discussed later in Section \ref{sec:achieve_fairness}.

\subsection{Revisiting Music School Admission Example}
In Section \ref{sec:motivating_example} we considered an empirical scenario of music school admission and presented a list of fairness inquiries one might be interested in.
When evaluating whether or not the admission is fair in general, there are additional technical inquiries, i.e, the ``algorithmic'' part of fairness considerations:
\begin{enumerate}[label=Question \arabic*, align=left]
    \setcounter{enumi}{5}
    \item\label{q:music_1} With respect to the data that the committee takes as a reference (which contains the admission choices of committees in previous years), is the data free from historical discrimination?
    \item\label{q:music_2} If we are willing to believe that the previous admission decisions do not contain any historical discrimination, based on the information at hand, then does the committee evaluate the qualification of applicants without bias (how the committee of this year evaluates the applicants)?
    \item\label{q:music_3} For those applicants who did not manage to get admitted this year, is there any difference in their future developments compared to those who got admitted?
        Is there any further impact on the representation of their ethnic groups in the violinist community?
\end{enumerate}
As we can see from these fairness inquiries, there are different underlying assumptions behind each question (e.g., the assumption that the previous admission results are free of historical discrimination), which determine the context and object of interest (e.g., the possible discrimination in admission results of previous years or this year specifically).
The nuances between various fairness inquiries actually reflect the necessity of disentangling different types of fairness concern and clarifying the tasks that are called for correspondingly.

\subsection{Algorithmic Fairness Spectra}
In light of the existence of various types of discrimination, the distinction between \textit{Without Disparate Impact} (also referred to as \textit{Outcome Fairness}) and \textit{Without Disparate Treatment} (also referred to as \textit{Procedural Fairness}) has already been proposed in Title VII of the 1964 Civil Rights Act.
While the procedural/outcome fairness division (or similarly, the \textit{Procedural} and \textit{Substantive} emphases presented in Section \ref{sec:philosophy}) indicates the intuition behind how different kinds of discrimination could occur, we believe that it is still preferable to have an overarching categorization of algorithmic fairness inquiries, namely, \textit{Fairness w.r.t. Data Generating Process}, \textit{Fairness w.r.t. Predicted Outcome}, and \textit{Fairness w.r.t. Induced Impact}.
By explicitly presenting the unstated or implicit assumptions, we further clarify the nuances between various types of fairness inquires, so that we can have a better understanding of the relative emphasis we should attribute to different algorithmic fairness spectra.

\subsubsection{\textbf{Fairness w.r.t. Data Generating Process}}
~ \\
The primary focus for this type of fairness inquiry is on the underlying data generating process.
Multiple factors may contribute to bias in the data \citep{danks2017algorithmic,mitchell2018prediction,mehrabi2021survey}: the imperfection of previous human decisions, the lingering effect of historical discriminations, the (potentially) morally neutral statistical bias/error in the sampling and measurement, and so on.

We say the data (i.e., the population) is ``clean'' as a consequence of data generating process satisfying the fairness notion of interest (e.g., a choice of the practitioner, or a prevailing conception of fairness).
The primary goal is therefore to quantify the discrimination with respect to the data itself, without considering downstream tasks like the prediction or decision making.
In the previous music school example, \ref{q:music_1} is a fairness inquiry with respect to the data generating process, inquiring the existence of discrimination in the data that results from the imperfection of previous committee decisions.

\subsubsection{\textbf{Fairness w.r.t. Predicted Outcome}}
~ \\
It is a common practice to evaluate the performance of machine learning algorithms by comparing the prediction with the ground truth in the data, which might be quite problematic if the data is already biased.
In light of this fact, whenever we utilize the data to train a ``fair'' prediction algorithm we actually take one thing for granted (or at least implicitly assumed)---the data itself is ``clean'' (according to the bias definition of interest).
As already pointed out in the literature \citep{kearns2019ethical}, there is, in general, no one-size-fits-all solution in terms of what fairness notion we should use.
Therefore we do not specify the exact definition of ``fair'' or ``clean'', and the aforementioned rationale applies to the fairness notion of interest in practical scenarios.

For \textit{Fairness w.r.t. Predicted Outcome}, we are not encouraging the practice of blindly assuming that the data at hand is unbiased.
Instead, we should always keep in mind that when we discuss fairness with respect to the prediction, there is an implicit assumption of ``clean'' data, which itself is subject to evaluations from the spectrum of \textit{Fairness w.r.t. Data Generating Process}.
By making the assumption that the data at hand is ``clean'', we can lift the burden from the downstream tasks, and emphasize the utilization of information such that fairness with respect to the predicted outcome is guaranteed.

Admittedly, there are different kinds of downstream tasks and not all of them can be solved by developing a predictive model.
Nevertheless, this category of fairness inquiry applies to predictive models as well as prediction-based decision-making systems.
After all, human decision making also rests upon predictions to some extent \citep{mitchell2018prediction}.
We use the name ``Fairness w.r.t. Predicted Outcome'' to further indicate the fact that the primary goal is to quantify the discrimination with respect to the prediction of the ground truth.
This does not exclude the possibility of considering downstream tasks like single-time or sequential decision making.
In the music school example, \ref{q:music_2} is a fairness inquiry with respect to the predicted outcome, focusing on the decision-making process of the committee under the assumption that the data (for both previous students and the applicants this year) itself is unbiased.

\subsubsection{\textbf{Fairness w.r.t. Induced Impact}}
~ \\
The fairness inquiry with respect to the induced impact is different from quantifying discrimination in the data or the predicted outcome.
Fairness inquiries in this spectrum focus on parties other than the prediction or decision making, for instance, how individuals could react (e.g., the interplay between the user and the system), how affirmative actions might help achieve fairness (e.g., the policy favor or investment to help the worse-off groups), and so on.
Essentially, the primary goal is to consider the possibility of characterizing fairness through the efforts of external entities besides prediction and decision makers.
As we will see in Section \ref{sec:achieve_fairness}, fairness inquiries can involve external entities, for example, user dynamics, data dynamics, and so on.
In the music school example, \ref{q:music_3} is a fairness inquiry with respect to the induced impact (of deploying a decision-making system).

\subsection{Remark: The Necessity of Considering Different Fairness Spectra}\label{sec:remark_multi_spectra}
In this section, we have seen different spectra of fairness inquiries.
Our goal is to provide a road map so that one can zoom in and see which part the current literature fit in and zoom out to see what else we can do with a clear target in mind.
Here, we present additional discussions in the form of questions and answers.

\subsubsection{\textbf{Why Distinguish Between Data and Prediction Fairness?}}
~ \\
To begin with, as we shall see in more detail in Section \ref{sec:achieve_fairness}, notions for \textit{Fairness w.r.t. Data Generating Process} are defined without reference to a predictor.
Auditing \textit{Fairness w.r.t. Data Generating Process} is irrelevant to what predictor one uses because the audit itself is with respect to (a sequence of) snapshots of reality.
This indicates that fairness endeavor with respect to data and that with respect to predicted outcome may well differ in terms of both technical definitions and objects of interest (e.g., $Y$ vs. $\Yhat$).

Besides, even if the data is ``clean'', the not-so-careful utilization of the data for prediction may still introduce new discriminations.
It is not necessarily the case that the prediction bias results only from data bias.
For instance, the unfairness can be introduced in prediction even if the label is fair \citep{ashurst2022fair}.

\color{RevisionText}  
Furthermore, there are attainability and optimality analyses with respect to the \textit{Fairness w.r.t. Predicted Outcome} notions themselves.
The attainability of prediction fairness, namely, the existence of a predictor that can score zero violation of fairness in the large sample limit, is an asymptotic property of the fairness notion \citep{tang2022attainability}.
Such attainability is not automatically guaranteed with clean data.
It characterizes a completely different kind of violation of fairness compared to the empirical error bound of discrimination in finite-sample cases.
In practice, although one can always audit violation of \textit{Fairness w.r.t. Predicted Outcome} via an empirical quantification, because of the finite sample size one cannot expect the empirical fairness violation to be exactly zero.
The absolute magnitude of the empirical fairness violation is often not informative enough since it is not clear how small an empirical fairness violation is small enough such that the predicted outcome can be deemed as ``fair'', i.e., the fairness notion of interest will be attained with zero violation in the large sample limit.
Therefore, it is desirable to develop prediction schemes that come with theoretical guarantees with respect to the method itself so that the fairness notion is proved to be attainable in the large sample limit.
Then we can further conduct optimality analysis among the models that have the attainability guarantee.
\color{black}  

Last, as we have seen in Section \ref{sec:dynamic_fairness}, ample evidence has suggested that ``fair'' predictions can have adverse impact on the fairness of data because of the driving force of involved dynamics.
There is no guarantee that the instantaneous rectification in the prediction/decision can somehow magically eliminate data bias.

Technically speaking, enforcing prediction fairness with or without (implicit or explicit) assumptions of clean data does not affect the algorithmic design or implementation.
Although $Y$ and $\Yhat$ are in essence both random variables, clearly distinguishing between fairness considerations for each one of them not only offers conceptual clarity but also provides a clearer picture regarding what kind of fairness inquiry one is actually conducting.

\subsubsection{\textbf{Is Fairness w.r.t. Induced Impact Redundant?}}
~ \\
While the difference between \textit{Fairness w.r.t. Data Generating Process} and \textit{Fairness w.r.t. Predicted Outcome} is relatively obvious, the distinction between \textit{Fairness w.r.t. Induced Impact} compared to the other two is more subtle.
We should not put \textit{Fairness w.r.t. Induced Impact} under the umbrella of either one of the other two categories.

To begin with, \textit{Fairness w.r.t. Induced Impact} itself does not necessarily assume that the data is unbiased (as does \textit{Fairness w.r.t. Data Generating Process}) or the utilization of information is not problematic (as does \textit{Fairness w.r.t. Predicted Outcome}).
Therefore if there is no guarantee regarding \textit{Fairness w.r.t. Data Generating Process} or \textit{Fairness w.r.t. Predicted Outcome}, the fairness violation may involve multiple parties including, but not limited to, the historical discrimination inherited from data, the reckless utilization of information in the prediction/decision-making process, and the interplay between the user and the system.

Furthermore, \textit{Fairness w.r.t. Data Generating Process} and \textit{Fairness w.r.t. Predicted Outcome} focus on either the data itself or the utilization of data, both of which are on the prediction/decision-making side; \textit{Fairness w.r.t. Induced Impact}, on the other hand, emphasizes the side of user autonomy and/or data dynamics as well as other possible external entities.
In our music school example, the difference in future developments may involve multiple parties, for instance, the committee (the decision maker), the background of the applicant (the user), the policy favor or educational investments for certain ethnic groups (the external entities), and the corresponding bias mitigation cannot be accomplished only through the effort of the music school committee.
\reBlue{}{
    We use the word ``party'' to indicate that different entities may contribute to the interplay.
    We reserve the word ``component'' to denote the causal mechanism in the data generating process.
}{0em}

\section{Subtlety: The Role of Causality in Fairness Analysis}\label{sec:subtlety_causality}
In Section \ref{sec:preliminaries} we presented multiple instantaneous fairness notions in the literature, many of which leverage the power of causal reasoning.
Before discussing the exact location where the notions might fit in the fairness spectra presented in Section \ref{sec:fairness_spectra}, we believe it is necessary and important to reflect on subtleties regarding the role of causality in fairness analysis.
The consideration of the subtleties motivates our (potential) modifications (in Section \ref{sec:achieve_fairness}) on previous fairness notions before applying them to fairness inquires from certain spectrum.
In particular, we argue that we should always perform sanity checks to make sure that we are quantifying discrimination in the way that matches underlying assumptions and intended types of fairness inquiries.

\subsection{Causal Modeling on the Object of Interest}\label{sec:causality_object_of_interest}
It is widely recognized in the fairness literature that we can leverage the power of causal reasoning to help us better understand how discrimination propagates through the data generating process \citep{kilbertus2017avoiding,kusner2017counterfactual,russell2017worlds,zhang2017causal,nabi2018fair,zhang2018fairness,loftus2018causal,chiappa2019path,wu2019pc}.
While the assumption of the availability of additional information about the data generating process, e.g., a causal graph, is in general acceptable, we find it questionable to directly assume that the prediction variable $\Yhat$ shares the exactly same causal graph with the ground truth variable $Y$.\footnote{
    For the tasks like prediction, the output $\Yhat$ is usually generated by a classification or regression algorithm in the literature.
}

\begin{figure*}[t]
    \centering
    \captionsetup{format=hang}
    \begin{subfigure}{.32\textwidth}
        \centering
        \includegraphics[width=.5\linewidth]{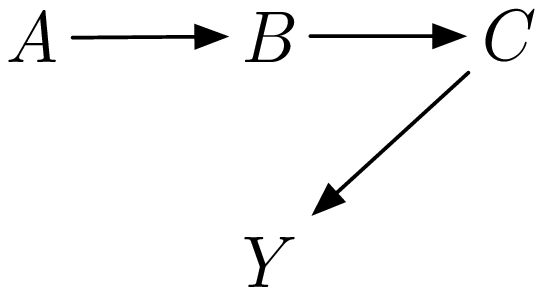}
        \caption{The data generating process \\
        for the ground truth $Y$}
        \label{fig:causal_graph_real}
    \end{subfigure}
    \begin{subfigure}{.32\textwidth}
        \centering
        \includegraphics[width=.5\linewidth]{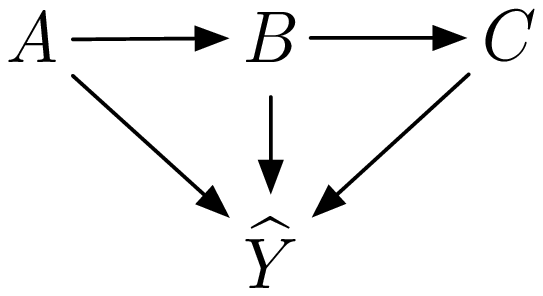}
        \caption{The data generating process \\
        for the classifier/regressor $\Yhat$}
        \label{fig:causal_graph_pred}
    \end{subfigure}
    \begin{subfigure}{.32\textwidth}
        \centering
        \includegraphics[width=.5\linewidth]{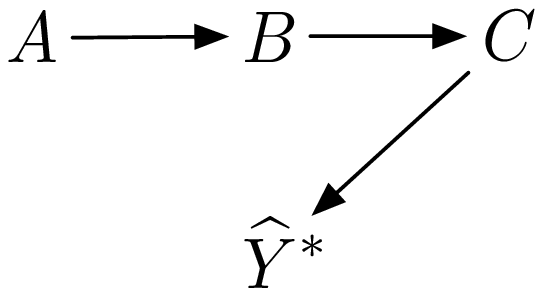}
        \caption{The data generating process \\
        for the inference output $\Yhat^*$}
        \label{fig:causal_graph_infer}
    \end{subfigure}
    \caption{
        The comparison between the causal graphs that represent different data generating processes for the ground truth $Y$, the prediction result via regression $\Yhat$, and the prediction result via inference $\Yhat^*$.
    }
    \label{fig:three_causal_graph}
\end{figure*}

Let us consider a simple example of the performance of basketball players where there are four variables: the gender of the player ($A$), the height of the player ($B$), the player's position ($C$), the total points scored by the player in this season ($Y$).
Suppose that the data generating process with respect to ground truth (the current reality), i.e., the relation among the measured variables $A$, $B$, $C$, and $Y$, can be described by Figure \ref{fig:causal_graph_real}:
the gender is a cause of the height; the height determines the position of the player on court; the position determines the total points that the player scores in the season.\footnote{
    This is a simplified model with a limited number of variables involved for illustrative purposes.
}
The task is to come up with the prediction ($\Yhat$) for the total points of this season ($Y$) based on the information available (the gender $A$, the height $B$, and the position $C$):
$\Yhat = f(A, B, C)$ where $f: \Acal \times \Bcal \times \Ccal \rightarrow \Ycal$ is a classification/regression algorithm.
In this case, the prediction result itself can be viewed as a random variable.
If we were to draw a graph that represents how $\Yhat$ is generated from $(A, B, C)$, we will have a data generating process as shown in Figure \ref{fig:causal_graph_pred}.
The reason for the extra arrows in Figure \ref{fig:causal_graph_pred} compared to Figure \ref{fig:causal_graph_real} is that the classification/regression algorithm, regardless of the loss function and the optimization techniques, treats available variables merely as input features, which does not really respect the original data generating process in Figure \ref{fig:causal_graph_real}.
However, if, for example we use a generative model and perform a probabilistic inference task on the outcome where we follow the underlying data generating process, then the inference result $\Yhat^*$ can share the causal graph with that of the ground-truth variable $Y$ (with a change of variable from $Y$ to $\Yhat^*$).
The data generating process for the prediction result via inference $\Yhat^*$ (Figure \ref{fig:causal_graph_infer}) is only different (in terms of the causal graph) from its counterpart for the ground truth variable $Y$ (Figure \ref{fig:causal_graph_real}) up to a substitution of the outcome variable.
Usually, we still need stronger assumptions regarding the underlying data generating process in order to perform the inference tasks, e.g., the availability of an SCM instead of only the causal graph.

As we can see in the previous example, when performing causal reasoning in fairness analysis, we should always be aware of the object of interest, i.e., the variable whose data generating process is subject to fairness consideration.
When we directly assume that the causal graph can be shared by the ground truth and the prediction, there could be a mismatch between the causal model (based on which the discrimination is quantified) and the object (whose data generating process is, in fact, \textit{not} described by this model).
If there is a mismatch between the causal model and the object of interest, then the result of discrimination quantification could be unpredictable and therefore is hardly justifiable.

\subsection{Causal Modeling with the Intended Interpretation}\label{sec:causality_intended_interpretation}
\color{RevisionText}  
As we have seen in Sections \ref{sec:notion_no_direct_indirect}-\ref{sec:notion_PC_CF}, it is a common practice in the causal fairness literature to first combine a causal graph with assumptions on functional forms of the SCM, and then perform fairness audit on the existence of certain causal effects.
However, when we are presented a causal model in fairness analysis, there are multiple interpretations that one could potentially apply to the causal model:
\begin{enumerate}[label=Interpretation \arabic*, align=left]
    \item\label{misc:interpret_discover} The causal model is \textit{recovered} from the data at hand through causal discovery (under some conditions);
    \item\label{misc:interpret_assume} The causal model is based on assumptions or background knowledge, according to which we \textit{believe} the data at hand is generated;
    \item\label{misc:interpret_ideal} The causal model reflects our \textit{expectation} that it should hold true in the hypothetical ideal world where there is no discrimination.
\end{enumerate}

\ref{misc:interpret_discover} and \ref{misc:interpret_assume} are of a similar flavor, characterizing the causal relations among (measured) variables in the current reality.
The corresponding data generating process only reflects the status quo, and the causal model itself does not provide any information regarding the existence of discrimination.
The existence of certain causal influence (e.g., in the form of a causal path) can be deemed as morally neutral or morally objectionable depending on the context of discussion as well as the algorithmic fairness definition.
For instance, for path-based algorithmic fairness definition (e.g., Definition \ref{def:no_unresolved_discrimination}, Definition \ref{def:no_proxy_discrimination}), the path $A \rightarrow B \rightarrow C$ in causal graph presented in Figure \ref{fig:causal_graph_real} may be morally neutral in the basketball player example (gender influences the height of the player, which in turn determines the court position of the player).
This path may be morally objectionable in a different context, for example, when $A$ represents individual's nationality, $B$ represents the favorite color, and $C$ represents the reckless driving habit.
Although the aforementioned two scenarios share the causal modeling in terms of the causal graph (Figure \ref{fig:causal_graph_real}), the existence of discrimination with respect to the causal path $A \rightarrow B \rightarrow C$ depends on the context and the definition of discrimination, both of which are not specified by the causal modeling itself under \ref{misc:interpret_discover} or \ref{misc:interpret_assume}.

\ref{misc:interpret_ideal}, however, interprets the model as the one that corresponds to the ideal fair world which may not be the case in the current reality.
In the basketball player example, for instance, the practitioner may determine (for some reason) that the causal influence from $B$ (the height of the player) to $C$ (the position on court) is morally objectionable and therefore unfair.
The practitioner argues that in the hypothetical world there should not be a path from $B$ to $C$.
Then, if the practitioner would like to see what would the distribution of $Y$ be in the hypothetical ideal world, it is no longer reasonable to refer to an SCM represented by Figure \ref{fig:causal_graph_real}, since there is a path $B \rightarrow C$.



\color{black}  

Among these various possible interpretations, it is not always self-explanatory from the fairness notions themselves which interpretation really corresponds to the causal model presented to us, if there is no further clarifications.
In practice, we should not only keep in mind the intuition behind the fairness notions, but also make sure that the interpretation of the causal model we are using truly matches the type of the intended task.

Therefore, categorizing a fairness notion in terms of the type of relation among variables it is defined with (e.g., the division between the associative and causal notions of fairness) may not be informative enough for us to guarantee fairness.
The neglect of the subtleties can easily disguise the existence of discrimination.
Actually as we shall see in Section \ref{sec:achieve_fairness}, the role of causality in fairness analysis is better represented by the insights it introduces into the problem, under the condition that we carefully perform the aforementioned sanity checks for the intended task.

\subsection{To Work \textit{against} or Work \textit{with} the Data Generating Process?}\label{sec:work_against_or_with}
\color{RevisionText}  
In this subsection, let us turn our focus to methodologies in causal fairness analysis.
On a high level, there are two different types of methodologies in terms of how one would like to treat data generating processes (with respect to data itself, or, how prediction is derived):
to work \textit{against} or to work \textit{with} the underlying data generating process.

To work \textit{against} the underlying data generating process, one identifies the unwanted causal paths \citep{kilbertus2017avoiding} or causal effects \citep{kusner2017counterfactual,chiappa2019path,wu2019pc} as the instantiation of discrimination, and would like to make sure that the prediction is not contaminated by the specified discrimination.
Current literature has witnessed various causal fairness notions that adopt the working-against methodology in instantaneous fairness analysis.
For example, Definitions \ref{def:no_direct_discrimination} and \ref{def:no_indirect_discrimination} advocate constraining direct or indirect (interventional) causal effects from the protected feature to the predicted outcome \citep{zhang2017causal,kilbertus2017avoiding,nabi2018fair};
Definition \ref{def:counterfactual_fairness} proposes eliminating counterfactual causal effects from the protected feature to the prediction \citep{kusner2017counterfactual};
Definition \ref{def:pc_fairness} characterizes more fine-grained versions of counterfactual causal effects and defines fairness through the nonexistence of such causal effects \citep{chiappa2019path,wu2019pc}.

We reflect on the methodology of working \textit{against} the data generating process.
To begin with, as presented in Sections \ref{sec:causality_object_of_interest}-\ref{sec:causality_intended_interpretation}, the causal modeling of the data generating process involves subtleties with respect to the object of interest and the intended interpretation.
If one neglects the subtleties when modeling the data generating process, then the causal analysis for fairness is hardly justifiable.

Besides, even if the aforementioned sanity checks are carefully performed and the causal modeling matches the data generating process of interest, it is not always the case that such data generating process is easily manipulable.
If we are enforcing causal fairness with respect to a predictor or decision-making system, then under certain technical conditions the fairness constraints can be implemented, because this specific data generating process is within the control of the algorithm or practitioner.
However, if we determine that the underlying data generating process for our current reality contains certain discrimination, then such process is not always within the control of a decision-making system.
For instance, social changes do not happen in an abrupt manner, and the fair solution is not simply removing an edge in the causal graph or performing certain interventions once and for all.
Furthermore, the expected change in the underlying data generating process often happens on the system level, e.g., the systematic oppression in terms of opportunity hoarding \citep{tilly1998durable}, instead of the model level, e.g., a model for automated decision making in the loan application \citep{liu2018delayed}.

Last, from a dynamic and long-term perspective, the enforcement of causal fairness in the working-against manner may have unintended downstream outcomes.
For instance, \citet{nilforoshan2022causal} consider \textit{Counterfactual Predictive Parity} \citep{coston2020counterfactual}, \textit{Counterfactual Equalized Odds} \citep{mishler2021fairness}, and \textit{Conditional Principal Fairness} \citep{imai2020principal} notions of causal fairness, and perform a one-step feedback analysis (a choice of analyzing framework reviewed in Section \ref{sec:choice_one_step}) in a simulated college admission scenario.
\citet{nilforoshan2022causal} conduct a \textit{Utilitarian} (Section \ref{sec:overarching_framework}) analysis and demonstrate the trade-offs between causal fairness notions and the downstream social welfare.

To work \textit{with} the data generating process, one recognizes the limited control over the underlying data generating process and focuses on the interplay between decision-making and data dynamics.
In light of this, it has been advocated in the recent literature to adopt the methodology of working \textit{with} the data generating process and explore the possibility of inducing a fairer future by analyzing the decision-distribution interplay \citep{tang2023tier}.

\color{black}  

\section{Enforcing Fairness in Different Spectra}\label{sec:achieve_fairness}
In Sections \ref{sec:fairness_spectra} and \ref{sec:subtlety_causality} we have seen different spectra of algorithmic fairness inquires and the subtleties of applying causal reasoning in fairness inquires, respectively.
In this section, we discuss ways to perform fairness audits and achieve algorithmic fairness in different spectra.

In Section \ref{sec:flowchart_intro}, we propose a flowchart corresponding to our fairness inquiry categorization.
Then, in Sections \ref{sec:achieve_procedure_fairness}-\ref{sec:achieve_induced_fairness}, we revisit commonly used fairness notions (reviewed in Section \ref{sec:fairness_notions}), with potential necessary modifications, illustrating how they fit in the fairness spectra (presented in Section \ref{sec:fairness_spectra}) so that the intuitive idea of fairness can better match the technical definition (which exact type of fairness we really would like to enforce).
In particular, for \textit{Fairness w.r.t. Data Generating Process}, the goal is to \textbf{detect} the discrimination embedded in the data;
for \textit{Fairness w.r.t. Predicted Outcome}, the goal is to \textbf{regulate} the way algorithms utilize information in the data (under the assumption that the data is ``clean'');
for \textit{Fairness w.r.t. Induced Impact}, the goal is to \textbf{compensate} the potential remaining inequalities from the effort of external entities, e.g., the user and/or data dynamics, so that fairness can be further improved.
In Section \ref{sec:closed_loop}, we provide a remark on the potential of performing the \textbf{correction} of discrimination-contaminated data through a closed-loop analysis across fairness spectra.

\begin{figure*}[tp]
    \centering
    \captionsetup{format=hang}
    \includegraphics[width=.95\linewidth]{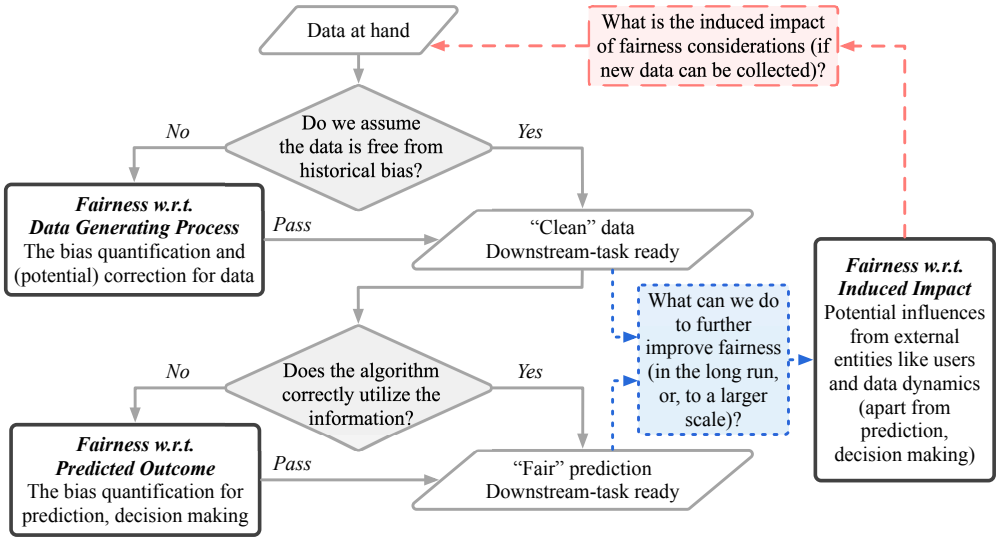}
    \caption{
        The flowchart that illustrates the road map to navigate through different spectra of fairness inquiries.
        Starting from the data at hand, based on the answer to the questions, we can sequentially audit fairness with respect to the underlying data generating process, the predicted outcome itself, and the induced impact in the future, respectively.
        If conditions permit, the newly collected data could be the starting point for a new round of fairness analysis.
        With a clear picture in mind that is able to accommodate different types of fairness inquiries, we can conduct algorithmic fairness analysis in a closed-loop manner, making the fairness analysis more principled and to-the-point.
    }
    \label{fig:flowchart}
\end{figure*}

\subsection{The Algorithmic Fairness Flowchart: Answering ``How-to'' Questions}\label{sec:flowchart_intro}
In Figure \ref{fig:flowchart} we present a road map to navigate through different fairness spectra.
Starting from the very beginning, the input for \textit{Fairness w.r.t. Data Generating Process} type of inquiries is the data at hand.
Depending on our answer to the question regarding whether or not the data itself is free from any historical discrimination, the data could be readily available for downstream tasks (if we answer ``yes'') or subject to bias quantification with potential correction (if we answer ``no'').

\textit{Fairness w.r.t. Predicted Outcome}, on the other hand, assumes that the data itself is ``clean'', i.e., the data passes the \textit{Fairness w.r.t. Data Generating Process} audits, and puts emphasis on the utilization of information to perform ``fair'' prediction/decision making.
Here ``clean'' and ``fair'' are always with respect to the fairness notions of interest, which largely remain choices of the practitioner.

For \textit{Fairness w.r.t. Induced Impact}, the input consists of the ``clean'' data and the ``fair'' prediction, and we consider the possibility of further improving fairness by taking into account the contribution from external entities other than the data and the prediction or decision maker (the blue dotted flow in Figure \ref{fig:flowchart}).
After going through the analysis through different types of fairness emphases, if conditions permit, then new data could be collected.
This in turn would be our updated version of the data at hand, which enables us to further check the effectiveness of the elimination of discrimination by a new round of fairness audit (the red dashed flow in Figure \ref{fig:flowchart}) and conduct a closed-loop fairness analysis.

\subsection{Fairness w.r.t. Data Generating Process}\label{sec:achieve_procedure_fairness}
A fairness inquiry from this category focuses on the generating process of the data itself and emphasizes the detection of discrimination within the data without considering downstream tasks.
In order to justify the way of discrimination quantification, we need to exploit the relation among measured variables in terms of the underlying data generating process, which makes causal modeling a perfect tool to achieve the goal.
In this subsection, we present our modifications on previously proposed causal notions of fairness, such that the modified notions are suitable for the purpose of auditing fairness with respect to the data generating process.

Multiple causal notions of fairness have been proposed in the literature \citep{zhang2016situation,kilbertus2017avoiding,kusner2017counterfactual,zhang2017causal,nabi2018fair,zhang2018fairness,khademi2019fairness,chiappa2019path,wu2019pc,salimi2019interventional}.
However, in light of the frequently neglected subtleties that we discussed in Section \ref{sec:subtlety_causality}, we might need to modify causal notions to remedy the mismatch between the intended task and the object or interpretation of interest, so that the intuition behind the notion can be properly expressed.
Here for the purpose of illustration, we present the modified versions of \textit{No Direct/Indirect Discrimination} (Definitions \ref{def:no_direct_discrimination} and \ref{def:no_indirect_discrimination}), \textit{Counterfactual Fairness} (Definition \ref{def:counterfactual_fairness}), and \textit{Path-specific Counterfactual Fairness} (Definition \ref{def:pc_fairness}) that we reviewed in Section \ref{sec:fairness_notions}.

\begin{definition}[No Direct Discrimination (Modified)]\label{def:no_direct_discrimination_mod}
    Given the causal graph that describes the data generating process of the current reality, let us denote as $\pi_d$ the path set that contains only the direct path from the protected feature $A$ to the outcome $Y$, i.e., $A \rightarrow Y$.
    We say that the outcome $Y$ is fair in terms of \textit{No Direct Discrimination} with respect to the protected feature $A$ and the path set $\pi_d$, if for any $a, a' \in \Acal$ and $y \in \Ycal$ the $\pi_d$-specific causal effect of the change in $A$ from $a$ to $a'$ on $Y = y$ satisfies:
    \begin{equation}
        P \big( Y = y \mid do(A = a'|_{\pi_d}) \big)
        - P \big( Y = y \mid do(A = a) \big) = 0.
    \end{equation}
\end{definition}

\begin{definition}[No Indirect Discrimination (Modified)]\label{def:no_indirect_discrimination_mod}
    Given the causal graph that describes the data generating process of the current reality, let us denote as $\pi_i$ the path set that contains all causal paths from the protected feature $A$ to the outcome $Y$ which go though redlining attributes $R$, i.e., each path within the set $\pi_i$ includes at least one node from $R$.
    We say that the outcome $Y$ is fair in terms of \textit{No Indirect Discrimination} with respect to the protected feature $A$ and the path set $\pi_i$, if for any $a, a' \in \Acal$ and $y \in \Ycal$ the $\pi_i$-specific causal effect of the change in $A$ from $a$ to $a'$ on $Y = y$ satisfies:
    \begin{equation}
        P \big( Y = y \mid do(A = a'|_{\pi_i}) \big)
        - P \big( Y = y \mid do(A = a) \big) = 0.
    \end{equation}
\end{definition}

\begin{definition}[Counterfactual Fairness (Modified)]\label{def:counterfactual_fairness_mod}
    Given a causal model $(U, V, \Fbf)$ that describes the data generating process of the current reality, where $V$ consists of all features $V \coloneqq \{A, X\}$, we say that the outcome $Y$ is fair in terms of \textit{Counterfactual Fairness} with respect to the protected feature $A$, if for any $a, a' \in \Acal, x \in \Xcal, y \in \Ycal$ the following holds true:
    \begin{equation}
        P \big( Y_{A \leftarrow a}(U) = y \mid A = a, X = x \big)
        = P \big( Y_{A \leftarrow a'}(U) = y \mid A = a, X = x \big).
    \end{equation}
\end{definition}

\begin{definition}[Path-specific Counterfactual Fairness (Modified)]\label{def:pc_fairness_mod}
    Given a causal model $(U, V, \Fbf)$ that describes the data generating process of the current reality and a factual observation $O = o$, where $V$ consists of all features $V \coloneqq \{A, X\}$ and $O \subseteq \{A, X, Y\}$, we say that the outcome $Y$ is fair in terms of \textit{Path-specific Counterfactual Fairness (PC Fairness)} with respect to the protected feature $A$ and the path set $\pi$, if for any $a, a' \in \Acal, y \in \Ycal$ the $\pi$-specific counterfactual causal effect of the change in $A$ from $a$ to $a'$ on $Y = y$ satisfies (let $\bar\pi$ denote the set containing all other paths in the graph that are not elements of $\pi$):
    \begin{equation}
        P \big( Y_{A \leftarrow a' | \pi, A \leftarrow a | \bar\pi}(U) = y \mid O = o \big)
        - P \big( Y_{A \leftarrow a}(U) = y \mid O = o \big) = 0.
    \end{equation}
\end{definition}

Compared to the original notions (Definitions \ref{def:no_direct_discrimination}, \ref{def:no_indirect_discrimination}, \ref{def:counterfactual_fairness}, \ref{def:pc_fairness}), the modified causal notions (Definitions \ref{def:no_direct_discrimination_mod}, \ref{def:no_indirect_discrimination_mod}, \ref{def:counterfactual_fairness_mod}, \ref{def:pc_fairness_mod}) are quantifying discrimination with respect to the outcome variable $Y$ instead of the prediction $\Yhat$, using the data generating process behind $Y$ in the current reality.
This seemingly trivial modification is more than just exchanges of variables.
In practical applications, when we assume the availability, either via an educated guess or from the expert knowledge, of a causal graph that characterizes underlying properties of the data, we are referring to the data generating process with respect to the outcome variable $Y$, instead of the predictor $\Yhat$ \citep{chiappa2018causal,nabi2018fair}.

Furthermore, even if we can draw the causal graph for predictions as illustrated in Figure \ref{fig:causal_graph_pred} (for prediction via classification/regression) and Figure \ref{fig:causal_graph_infer} (for prediction via inference), we will still need to make sure that we pair up the object of interest and the technical detail of the corresponding analyzing scheme (e.g., path-based criterion, or causal effect estimation that involves additional information/assumption on the functional class).

Let us revisit the basketball player performance example in Section \ref{sec:subtlety_causality}.
Suppose that a practitioner would like to audit fairness with respect to the prediction and at the same time understand the source of discrimination, and the practitioner thinks that a causal notion of fairness could be very handy.
Suppose, for example, the practitioner picks \textit{Counterfactual Fairness} (Definition \ref{def:counterfactual_fairness}, which is the original notion proposed by \citet{kusner2017counterfactual}) since this causal notion is with respect to $\Yhat$.
There are multiple strategies a practitioner might choose to audit fairness, and for each one of them it is possible to have a mismatch between the mission (which kind of fairness we really would like to capture) and the means (how exactly the fairness audit is carried out):
\reBlue{}{
    We discuss four different strategies to clearly show how the unmodified causal fairness notion can invite potential negligence (discussed in Section \ref{sec:subtlety_causality}).
    We would like to point out that causal fairness notions, with proper modifications, better serve the purpose of auditing \textit{Fairness w.r.t. Data Generating Process} (lines 1311 - 1316).
    Therefore, the revisit of the basketball player example fits in Section \ref{sec:achieve_procedure_fairness}.
}{0em}

\begin{figure*}[t]
    \centering
    \captionsetup{format=hang}
    \begin{subfigure}{.32\textwidth}
        \centering
        \includegraphics[width=.5\linewidth]{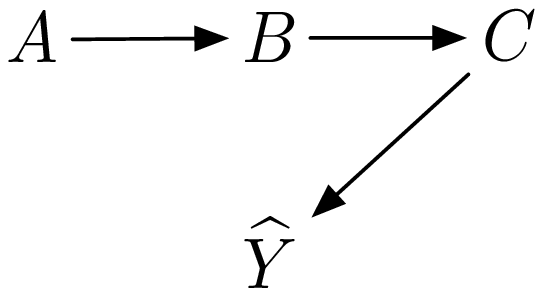}
        \caption{Strategy 1}
        \label{fig:practitioner_graph_1_2}
    \end{subfigure}
    \begin{subfigure}{.32\textwidth}
        \centering
        \includegraphics[width=.5\linewidth]{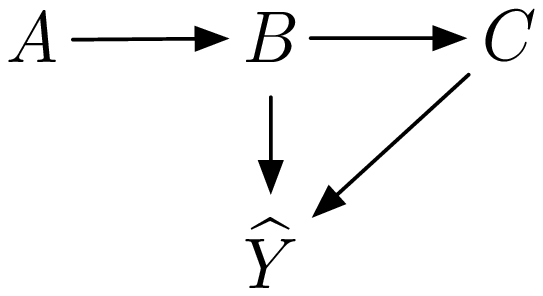}
        \caption{Strategy 3}
        \label{fig:practitioner_graph_3}
    \end{subfigure}
    \begin{subfigure}{.32\textwidth}
        \centering
        \includegraphics[width=.5\linewidth]{figure/causal_graph_pred.eps}
        \caption{Strategy 4}
        \label{fig:practitioner_graph_4}
    \end{subfigure}
    \caption{
        The comparison between graphs that the practitioner draws in different strategies.
    }
    \label{fig:practitioner_graphs}
\end{figure*}

\begin{enumerate}[label=Strategy \arabic*, align=left]
    \item\label{misc:strategy_1} The practitioner makes an educated guess regarding how attributes could relate to each other in the data set and draws the causal graph Figure \ref{fig:causal_graph_real}.
    Considering the task is to audit fairness on $\Yhat$, the practitioner directly exchanges the variable $Y$ in the graph to $\Yhat$ and draws the graph shown in Figure \ref{fig:practitioner_graph_1_2}.
    The practitioner then proceeds to the fairness audit via \textit{Counterfactual Fairness} (Definition \ref{def:counterfactual_fairness}) without knowing the detail regarding how $\Yhat$ is computed (which is the regression output).\label{pts:strategy_1}
    \item\label{misc:strategy_2} The practitioner utilizes the exactly same strategy to audit fairness as in \ref{pts:strategy_1}, without knowing the detail regarding how $\Yhat$ is computed (which is in fact output of a inference model shown as in Figure \ref{fig:causal_graph_infer}).
    \item\label{misc:strategy_3} The practitioner first pictures an idealized fair world where both the height $B$ and the position $C$ are causes of the total points $Y$ scored by the player.
    Then the practitioner realizes that the task is to audit fairness on $\Yhat$ and draws the graph shown in Figure \ref{fig:practitioner_graph_3}.
    The practitioner proceeds to the fairness audit via \textit{Counterfactual Fairness} (Definition \ref{def:counterfactual_fairness}) without knowing the detail regarding how $\Yhat$ is computed (which is the regression output).
    \item\label{misc:strategy_4} The practitioner notices that $\Yhat$ is the output of a regression algorithm and draws the causal graph that corresponds to the data generating process of $\Yhat$ as shown in Figure \ref{fig:practitioner_graph_4}.
    The practitioner then proceeds to the fairness audit via \textit{Counterfactual Fairness} (Definition \ref{def:counterfactual_fairness}) with respect to $\Yhat$.
\end{enumerate}

Let us take a closer look at these different strategies.
For \ref{misc:strategy_1}, there is a mismatch between the object of interest ($\Yhat$) and the corresponding data generating process (it should be the graph shown in Figure \ref{fig:causal_graph_pred}, instead of Figure \ref{fig:practitioner_graph_1_2}).

For \ref{misc:strategy_2}, there seems to be no mismatch between the object of interest ($\Yhat$) and the corresponding data generating process in terms of the causal graph, since Figure \ref{fig:practitioner_graph_1_2} happens to be identical to Figure \ref{fig:causal_graph_infer} (except for the asterisk symbol in Figure \ref{fig:causal_graph_infer}).
Although the causal graphs agree with each other, the details of causal modeling (e.g., functional classes in the SCM) may differ across the algorithm builder (who generates $\Yhat$) and the practitioner (who audits fairness on $\Yhat$), which may still incur a mismatch between the object of interest and the corresponding data generating process.

For \ref{misc:strategy_3}, there is a mismatch of the causal modeling both in terms of the intended interpretation (using the graph which reflects the hypothetical ideal world) and the object of interest (substituting $Y$ with $\Yhat$ without justification).

For \ref{misc:strategy_4}, there seems to be no mismatch since Figure \ref{fig:practitioner_graph_4} is identical to Figure \ref{fig:causal_graph_pred}.
However, while there is no significant difference in terms of technical treatments when estimating causal effects on $Y$ and $\Yhat$ (if we were to draw a causal graph for the regression output), only the data generating process behind $Y$ reflects what happens in the real world.
After all, one of the strongest motivations behind the usage of a causal notion is the insight into the data generating process behind the outcome $Y$ in the current reality, but this purpose does not seem to be well-served if we consider the data generating process behind the prediction $\Yhat$.

As we can see from different possible strategies in this example, there are many subtleties involved in enforcing/auditing causal notions of fairness.
Neglecting these subtleties may result in mismatches between the mission and the means.
Unfortunately, the precautions against these negligence are often not well packed into the causal notions of fairness themselves in the current literature.
To some extent, the causal notions of fairness with respect to $\Yhat$ (unintentionally) invites the negligence of subtleties discussed in Section \ref{sec:subtlety_causality}.

In fact, it is not uncommon to see (variants of) the aforementioned \ref{pts:strategy_1} utilized in current literature \citep{kilbertus2017avoiding,kusner2017counterfactual,zhang2017causal,wu2019pc}.
Therefore our modification on causal notions of fairness is necessary and important to make sure that the notions are correctly used for the suitable task---to \textbf{detect} discrimination within the current data and audit \textit{Fairness w.r.t. Data Generating Process}.

Admittedly, the detection of the existence of discrimination in the data does not easily translate into possible ways to perform correction.
Nevertheless, a sensible and justifiable scheme that fully characterizes our intuitions behind fairness considerations would encourage further explorations to better accomplish the task, and therefore, is always desirable.
We provide the discussion regarding the potential to correct the data via a closed-loop analysis in Section \ref{sec:closed_loop}.

\subsection{Fairness w.r.t. Predicted Outcome}\label{sec:achieve_prediction_fairness}
While various fairness notions proposed in the literature are with respect to the prediction $\Yhat$, as discussed in Section \ref{sec:achieve_procedure_fairness} not all of them are suitable for the intended fairness audit at hand.
Different from \textit{Fairness w.r.t. Data Generating Process} where the goal is to detect the discrimination within data, \textit{Fairness w.r.t. Predicted Outcome} assumes that the data at hand is free from discrimination (in the sense that the data passes the fairness audit from the \textit{Fairness w.r.t. Data Generating Process} category) and \textbf{regulates} the utilization of information when performing predictions.
In practical scenarios, the prediction is often performed by a classification or regression algorithm, which would only treat available features as input, regardless of the data generating process underlying the real world.
Therefore as a rule of thumb, for \textit{Fairness w.r.t. Predicted Outcome}, associative notions of fairness, e.g., \textit{Individual Fairness} \citep{dwork2012fairness}, \textit{Demographic Parity} \citep{calders2009building}, \textit{Equalized Odds} \citep{hardt2016equality}, are most suitable for the intended fairness audits in this category.

In the algorithmic fairness literature, the phenomenon of the ``trade-off between fairness and accuracy'' for the prediction has been widely observed and discussed \citep{kamiran2012data,romei2014multidisciplinary,feldman2015certifying,chouldechova2017fair,berk2017convex,corbett2017algorithmic,kleinberg2017inherent,menon2018cost,agarwal2018reductions,mary2019fairness,wick2019unlocking,baharlouei2020renyi,pinzon2022impossibility}.
However, as is discussed in Section \ref{sec:fairness_spectra}, only when we assume/know that the data does not contain discrimination can we really justify the practice of enforcing fairness and accuracy at the same time for the prediction result.
After all, if $Y$ contains discrimination, enforcing the prediction $\Yhat$ to be close to $Y$ (even if with fairness regularization) is not desirable.
Therefore for \textit{Fairness w.r.t. Predicted Outcome}, we would like to explicitly assume that the data itself is clean so that we can focus on the utilization of information.

\subsection{Fairness w.r.t. Induced Impact}\label{sec:achieve_induced_fairness}
In Section \ref{sec:fairness_spectra} we discussed the difference between \textit{Fairness w.r.t. Induced Impact} and other fairness spectra, i.e., \textit{Fairness w.r.t. Data Generating Process} and \textit{Fairness w.r.t. Predicted Outcome}.
In this section, we argue that we can explore the possibility of further improving fairness through the effort of external entities.

As we have discussed in Section \ref{sec:fairness_spectra}, we cannot put the \textit{Fairness w.r.t. Induced Impact} inquires under the umbrella of \textit{Fairness w.r.t. Data Generating Process} or \textit{Fairness w.r.t. Predicted Outcome} categories.
In light of the practical interpretation of \textit{Fairness w.r.t. Induced Impact} audits, we can go beyond the prediction/decision-making itself and explore the possibility of leveraging the effort of external entities to further improve fairness.

Furthermore, if we observe a shared issue among various prediction/decision-making cases, e.g., the recourse cost for certain group is always higher than others for both loan application and school admission, then this may indicate the disadvantage suffered by the group at a larger scale.
This disadvantage may be better \textbf{compensated} by (global) policy supports (e.g., investments in education for certain community to improve the overall socioeconomic status in the long run) compared to (localized) separated efforts from prediction/decision-making in different scenarios.
Here by ``global'' and ``localized'' we are referring to the scope of effectiveness (e.g., the \textit{Local} and \textit{Global} views presented in Section \ref{sec:local_global}): a policy support can potentially be effective in multiple prediction/decision-making scenarios, while prediction/decision-making itself is usually limited to the specific task at hand, i.e., the scenario for which the algorithm is implemented, like loan application or school admission.

Some might argue that the \textit{Fairness w.r.t. Induced Impact} task sounds like \textit{Fairness w.r.t. Data Generating Process} since we are characterizing historical discrimination in some sense.
While \textit{Fairness w.r.t. Data Generating Process} specializes in detecting discrimination (with potential correction) within the data, the scope is limited to measured variables in the data set at hand.
The long-term influence on latent attributes, e.g., the unobserved socio-economic status of individuals, are often not readily available for us when we audit fairness with respect to the current reality.
One might need to model the decision-distribution interplay and consider the behind-the-scenes situation changes on the unobserved latent causal factors that directly carry out the influence from the current decision to the future data distribution \citep{tang2023tier}.

Some might also argue that the \textit{Fairness w.r.t. Induced Impact} can be enforced in the same way as \textit{Fairness w.r.t. Predicted Outcome} by regulating the utilization of information in the prediction.
While it is a reasonable proposal, the focus of the \textit{Fairness w.r.t. Induced Impact} category often involves multiple parties including, but not limited to, the prediction/decision-making, the user dynamics, the external incentives (like affirmative actions).
The interplay between these stakeholders cannot be simplified into the analysis on the prediction/decision-making itself and we need to model dynamics for each party separately \citep{liu2018delayed,heidari2019long,zhang2020fair,tang2023tier}.
\textcolor{RevisionText}{
    The complexity of the practical implication of predicted outcome at a larger spatial or temporal scale also indicates the necessity of interpreting fairness robustness and fairness transferability in terms of not only the predicted outcome itself, but also the induced impact \citep{cotter2019training,chuang2021fair,liu2021induced,chen2022fairness,ferry2022improving}.
}
\rePurple{}{
    We think the discussion of fairness robustness fits in the spectrum of \textit{Fairness w.r.t. Induced Impact}.
    In light of the suggestion, we have incorporated a quick pointer to the related references.
}{0em}

\subsection{Remark: Closed-loop Algorithmic Fairness Analysis}\label{sec:closed_loop}
As we have seen in Section \ref{sec:dynamic_fairness}, current dynamic fairness studies already indicate the importance of considering induced impact of predictions/decisions.
We argue that the benefit of considering different spectra of fairness inquires can be extended to go beyond merely auditing the existence of bias but also correcting bias in the data.

The road map we presented earlier (Figure \ref{fig:flowchart}) is intended to enable a closed-loop fairness analysis by navigating through different spectra of algorithmic fairness inquiries.
We do not intend to claim that one can only consider the current fairness endeavor under the condition that the previous step in the flowchart is already satisfied.
Instead, we provide a guiding framework so that fairness analysis can follow a principled navigation.
For example, a prominent goal of algorithmic fairness inquiries is to make sure the historical bias is eliminated in the future.
In order to achieve this goal, it is not fruitful to consider prediction fairness in a static setting and hope that the prediction will somehow magically eliminate the bias embedded in data itself.
Since the underlying data generating process is the object of interest (\textit{Fairness w.r.t. Data Generating Process}), and the prediction/decision making itself does not offer a direct answer regarding how we can manipulate the underlying data generating process, we should instead follow the flowchart (Figure \ref{fig:flowchart}) and explore the possibility of inducing a fair data generating process in the future by conducting a closed-loop fairness analysis and considering \textit{Fairness w.r.t. Predicted Outcome} and \textit{Fairness w.r.t. Induced Impact} at the same time.

\section{Conclusion}\label{sec:conclusion}
In this paper, we provide a survey of, a reflection on, and a new perspective for fairness in machine learning.
In particular, we propose a framework that consists of fairness considerations from different perspectives, namely, data generating process, predicted outcome, and induced impact, and provide a road map, along with sanity checks, to navigate through different fairness spectra.

For fairness with respect to data generating process, considering the often neglected subtleties regarding the role played by causality in fairness analysis, we propose necessary modifications to previous causal notions of fairness and discuss the goal of detecting the discrimination within the data.
For fairness with respect to predicted outcome, we highlight the importance of clarifying assumptions on the data, as well as the often-overlooked attainability of fairness notions.
For fairness with respect to induced impact, we aim to explore the possibility of further improving fairness through the effort of external entities beyond prediction/decision-making.

Future research directions naturally span different spectra of  fairness we laid out.
For fairness with respect to data generating process, it is desirable to develop methods to evaluate and guarantee the effectiveness of the pursuit of fairness with respect to the underlying data generating process.
This is especially important for the potential correction, i.e., going beyond detection, of the discriminations within the data in the long-term, dynamic setting.
For fairness with respect to predicted outcome, a thorough understanding of the fairness notion of interest (e.g., the one that is, or will be, deployed in the real world) calls for analysis with respect to attainability and optimality, which, if carefully characterized, is very informative and helpful both in terms of theoretical rigorousness and practical significance (e.g., the development of better learning strategies that come with theoretical guarantees).
For fairness with respect to induced impact, the potential unification of the findings from fairness audits conducted in separated but highly-related scenarios (e.g., school admission, loan application, occupational outlook, etc.) would be very helpful to identify potential ways to systematically promote fairness from a wider scope.

The flowchart we propose (Figure \ref{fig:flowchart}) also highlights the potential to quantify the effectiveness of fairness endeavor of the current iteration through another round of fairness audits (e.g., the red dashed flow in Figure \ref{fig:flowchart}).
With meaningful interpretations of the result, the findings from multiple fairness spectra across different rounds of fairness audits would be a very informative guidance (for prediction/decision-making systems, as well as policy designers and lawmakers) to achieve fairness in an organized and principled way, which is of great theoretical and practical significance.



\bibliographystyle{ACM-Reference-Format}
\bibliography{references}

\appendix

\end{document}